%% file: main.tex
\documentclass{article} 
\usepackage{iclr2026_conference,times}

\PassOptionsToPackage{numbers, compress}{natbib}
\usepackage{microtype}
\usepackage{graphicx}
\usepackage{subfig}

\usepackage{multirow}
\usepackage{booktabs} 
\usepackage[utf8]{inputenc} 
\usepackage[T1]{fontenc}    
\usepackage{url}            
\usepackage{booktabs}       
\usepackage{amsfonts}       
\usepackage{nicefrac}       
\usepackage{xcolor}         
\usepackage{wrapfig}
\usepackage{adjustbox}
\usepackage{amsmath,amssymb,amsthm} 
\usepackage{tabularx}
\usepackage{float}
\usepackage{comment}
\usepackage{xspace}
\usepackage{subcaption} 

\usepackage[backref=page]{hyperref}

\input{math_commands.tex}

\usepackage[capitalize,noabbrev]{cleveref}

\theoremstyle{plain}
\newtheorem{theorem}{Theorem}[section]
\newtheorem{proposition}[theorem]{Proposition}

\newtheorem{corollary}[theorem]{Corollary}
\theoremstyle{definition}
\newtheorem{definition}[theorem]{Definition}

\theoremstyle{remark}

\newcommand{\mulos}{\mu\text{LO}_S }
\newcommand{\mulom}{\mu\text{LO}_M }
\newcommand{\muvelom}{\mu\text{VeLO}_M }
\newcommand{\los}{\text{LO}_S}
\newcommand{\lom}{\text{LO}_M}
\newcommand{\velom}{\text{VeLO}_M}

\newcommand{\mup}{$\mu$P\xspace}

\newcommand{\tmlp}{\gT^\text{MLP} }
\newcommand{\tvit}{\gT^\text{ViT} }
\newcommand{\tlm}{\gT^\text{LM} }

\newcommand{\commentall}[1]{}

\title{$\mu$LO: Compute-Efficient Meta-Generalization of Learned Optimizers}

\author{\parbox{\textwidth}{\centering
\vspace{0.75cm}
    Benjamin Thérien$^{1, 2}$ \hspace{-10pt}
    \qquad  Charles-Étienne Joseph$^{1, 2}$ \hspace{-10pt}
    \qquad Boris Knyazev$^{1,2,4}$ \hspace{-10pt}\\
    Edouard Oyallon$^{5}$
    \qquad Irina Rish$^{1, 2}$
    \qquad Eugene Belilovsky$^{2, 3}$ \vspace{5pt}\\
    \textnormal{{$^1$Université de Montréal; $^2$Mila -- Quebec AI Institute; $^3$Concordia University, Montréal;\\ $^4$Samsung AI Lab, Montréal; $^5$ISIR, Sorbonne University, CNRS, Paris, France. }}
}}

\iclrfinalcopy 

\begin{document}

\makeatletter
\def\blfootnote{\xdef\@thefnmark{}\@footnotetext}
\makeatother

\maketitle

\blfootnote{Correspondence to: Benjamin Thérien $\langle$benjamin.therien@umontreal.ca$\rangle$ and Eugene Belilovsky $\langle$eugene.belilovsky@concordia.ca$\rangle$. Our code is open-sourced: \scriptsize\url{https://github.com/bentherien/mu_learned_optimization}.}

\begin{abstract}
Learned optimizers (LOs) have the potential to significantly reduce the wall-clock training time of neural networks. However, they can struggle to optimize unseen tasks (\emph{meta-generalize}), especially when training networks wider than those seen during meta-training. To address this, we derive the Maximal Update Parametrization ($\mu$P) for two state-of-the-art learned optimizer architectures and propose a simple meta-training recipe for $\mu$-parameterized LOs ($\mu$LOs). Our empirical evaluation demonstrates that LOs meta-trained with our recipe substantially improve meta-generalization to wider unseen tasks when compared to LOs trained under standard parametrization (SP) using the same compute budget. We also empirically observe that $\mu$LOs exhibit unexpectedly improved meta-generalization to deeper networks ($5\times$ meta-training) and surprising generalization to much longer training horizons ($25\times$ meta-training) when compared to SP LOs. 
\end{abstract}
\vspace{-7pt}
\section{Introduction}
\vspace{-3pt}
\label{sec:intro}

While deep learning (DL) has largely replaced hand-designed algorithms, one crucial component of DL training remains hand-crafted: gradient-based optimizers. While popular optimizers such as Adam or SGD provably converge to a local minimum in non-convex settings~\citep{kingma2017adam,li2023convergence,Robbins1951ASA},  the existing literature provides no evidence that these optimizers converge to the global optimum at the optimal rate. 
With the lack of theory certifying the optimality of existing optimizers and the clear strength of data-driven methods, it is natural to turn towards data-driven solutions for improving the optimization of neural networks.


\begin{figure}[t]
    \vspace{-10pt}
    \subfloat[Axes of Meta-Generalization]{\includegraphics[width=0.49\linewidth]{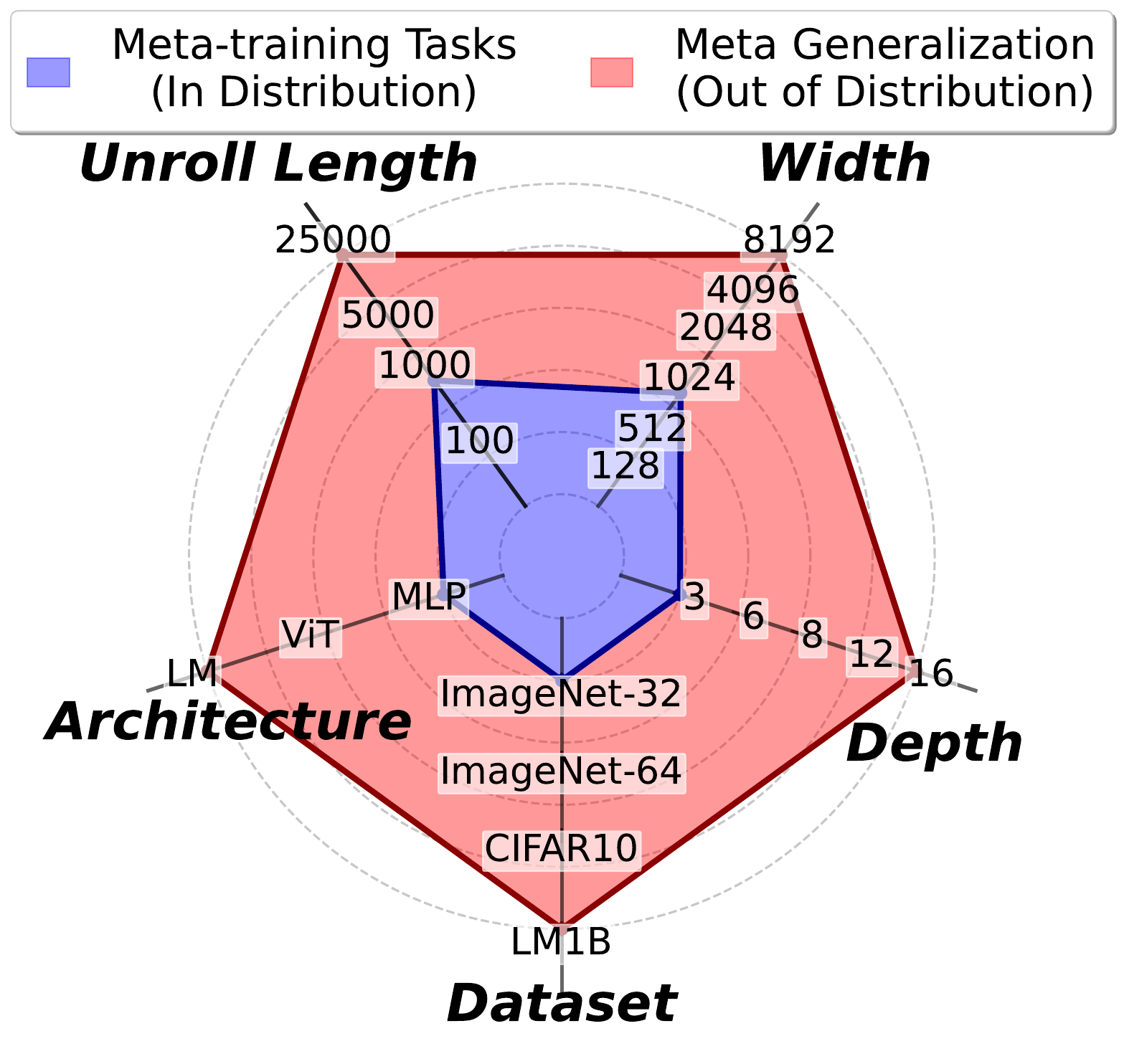}}
     \qquad
    \subfloat[Performance by Average Rank]{\includegraphics[width=0.456\linewidth]{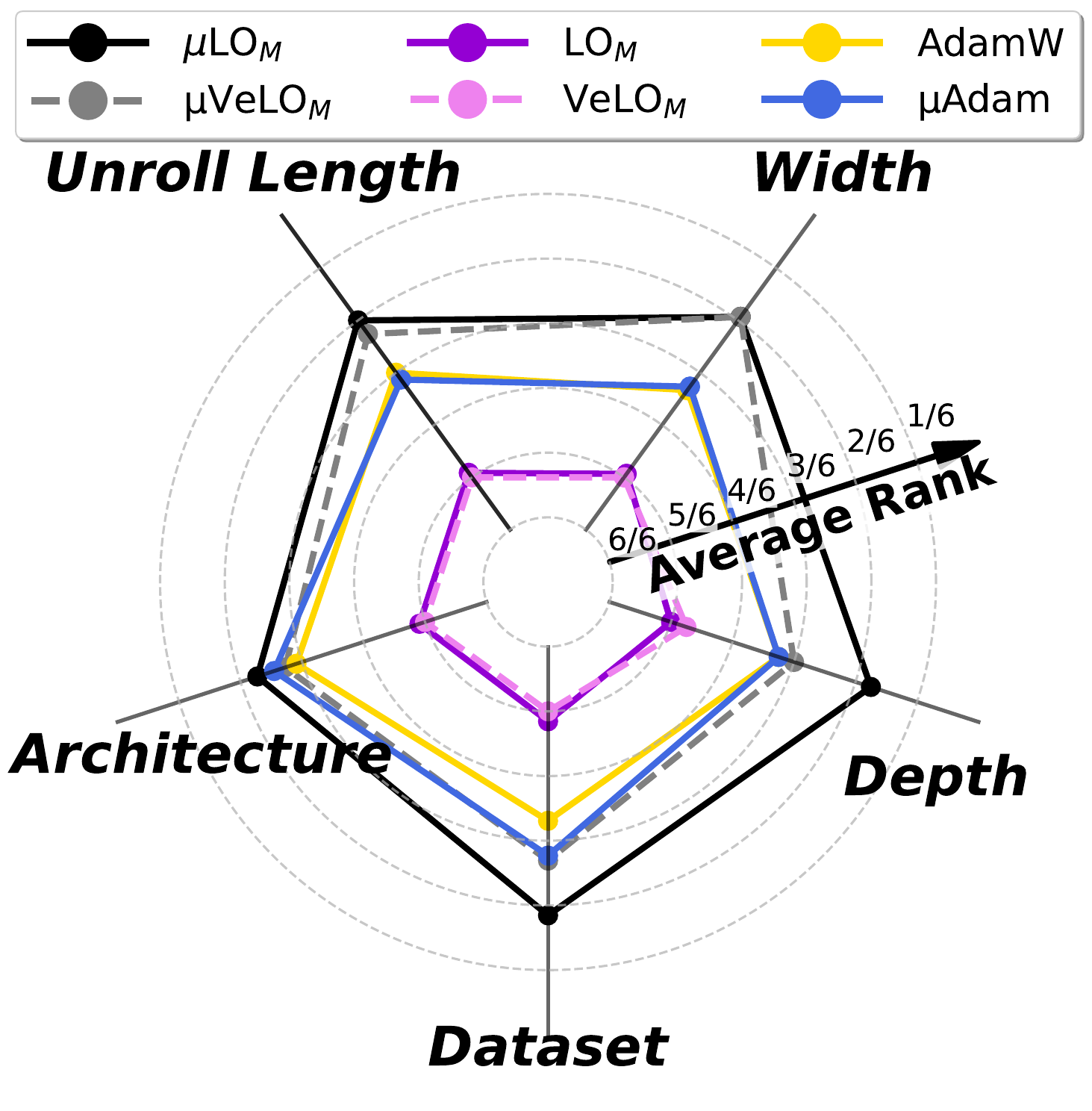}}
    \centering
    \vspace{-4pt}
    \caption{\textbf{Meta-generalization is severely limited without our approach.} Subfigure (a) illustrates \textit{meta-generalization} axes by distinguishing between meta-training tasks used herein (blue) and out-of-distribution tasks (red). Subfigure (b) reports the average rank across tasks within our evaluation suite that are out-of-distribution with respect to the corresponding axis. Both AdamW and $\mu$Adam undergo task-specific hyperparameter tuning across more than $500$ configurations per task. Learned Optimizers of the same architecture are meta-learned on the same tasks with a FLOP-matched budget. }
    \vspace{-12pt}
    \label{fig:metagen}
\end{figure}

Taking a step in this direction, \cite{andrychowicz2016learning,wichrowska2017learned,metz2019understanding,metz2022practical} replace hand-designed optimizers with small neural networks called learned optimizers (LOs). LOs are meta-learned on a task distribution by minimizing the loss of the inner learning problem (e.g. neural network training in our case) across a batch of tasks. Being neural networks themselves, these optimizers are advantaged by their substantially larger parameter counts than Adam or SGD, making them suitable to large-scale meta-training. For instance, \citet{velo} showed that scaling up learned optimizer meta-training to $4000$ TPU months can produce an optimizer, VeLO, that significantly outperforms well-tuned hand-designed optimizers without requiring hyperparameter tuning. However, even VeLO has limitations in \textit{meta-generalization} -- optimizing unseen problems. Specifically, VeLO~\citep{velo} is known to (1) have difficulty optimizing models much wider and deeper than those seen during meta-training (See Figures $6$ and $9$ of~\citet{velo}) and (2) generalize poorly to longer optimization problems (e.g., training for more steps) than those seen during meta-training.

The problem of \textbf{meta-generalization} is fundamental to learned optimization due to the requirement for tractable meta-training and the expectation of strong performance across a combinatorially large set of downstream tasks. Meta-generalization refers to the ability of a meta-learned algorithm to \emph{generalize}, that is, perform well when applied to unseen tasks. In the case of LOs, a learned optimizer trained on a tractable and, thus, limited distribution of meta-training tasks should nevertheless exhibit strong performance when applied to out-of-distribution tasks: new combinations of architecture, dataset, and training objective (Figure~\ref{fig:metagen}). Even changes as small as increasing the hidden dimension of the architecture (width), the number of layers (depth), or the number of training steps (unroll length) can cause meaningful distribution shifts between meta-training and testing tasks, leading to poor generalization. Consequently, understanding and improving meta-generalization is central to making learned optimizers practical for real-world machine learning workloads.

In this work, we focus on the problem of LO meta-generalization to tasks of larger hidden dimension (width) than those seen during meta-learning. A related problem is that of transferring hyperparameters of hand-designed optimizers to wider tasks. Introduced by~\cite{tensorv}, \mup is an optimizer-dependent and width-dependent parameterization (e.g., a rule for initializing a model, scaling its pre-activations, and scaling the optimizer's updates) that allows hyperparameter transfer to larger width tasks for Adam and SGD. Making the connection between hyperparameter-transfer and meta-generalization, we ask: \emph{Are existing learned optimizer architectures compatible with \mup? Does meta-learning optimizers under \mup improve meta-generalization?} To answer this question, we theoretically analyze two recent LO architectures~\citep{metz2022practical,velo} (sec.~\ref{sec:proof}), derive the appropriate maximal update parameterization for them, and carefully design a low-cost meta-training recipe to bring out their meta-generalization capabilities. We then provide extensive experimental evidence demonstrating that $\mu$LOs generalize to large unseen tasks.
Our contributions are as follows:

\begin{itemize}
    \item We derive $\mu$-parameterization for two popular learned optimizer architectures (VeLO and small\_fc\_lopt) and demonstrate theoretically that our parameterization satisfies $\mu$P desiderata.
    \item We design a set of meta-training and meta-testing tasks enabling a systematic study of meta-generalization and demonstrate that our $\mu$LOs significantly outperform strong baseline LOs and hand-designed optimizers.
    \item We demonstrate empirically that our $\mu$LOs show surprisingly good generalization to deeper networks ($5\times$ meta-training) and longer training horizons ($25\times$ meta-training) when compared to baseline LOs.
\end{itemize}


\vspace{-10pt}
\section{Background}\label{sec:background}

\paragraph{Learned optimizer objective.}A standard approach to learning optimizers \citep{metz2019understanding} is to solve the following meta-learning problem:
\begin{equation}\label{eq:obj}
\min_\phi\;\E_{(\gD,\gL,\vw_0)\sim \gT}\left[\E_{(X,Y)\sim \gD} \left[ \frac{1}{T}\sum_{t=0}^{T-1} \gL(X,Y;f_{\phi}(\vu_t),\vw_t) \right]\right].
\end{equation}
\noindent Where $\gT$ is a distribution over optimization tasks defined as tuples of dataset $\gD$, objective function $\gL$, and initial weights $\vw_0$ associated with a particular neural architecture (we refer to this network as the \emph{optimizee}); $\phi$ represents the weights of the learned optimizer,$f_\phi$ with input features $\vu_t$; and $T$ is the length of the unroll which we write as a fixed quantity for simplicity. In \eqref{eq:obj} and in our experiments, the sum of per-timestep loss is the quantity being optimized. It should be noted, however, that one could also optimize the final loss, final accuracy, or any other performance metric. Gradient descent is the preferred approach to solving \eqref{eq:obj}. However, estimating meta-gradients via backpropagation is known to be problematic for long unrolls~\citep{metz2019understanding}. Therefore, learned optimizer meta-gradients are estimated using evolution strategies and their variants~\citep{vicol2021pes,buckman2018es,nesteroves,parmas2018es,vicol2023low,li2023variance}.

\vspace{-7pt}
\paragraph{Learned optimizer input, output, and update.} Learned optimizer neural architectures have taken many forms over the years, we will briefly review two recent architectures, $\textbf{small\_fc\_lopt}$ of \citet{metz2022practical} and $\textbf{VeLO}$ of \citet{velo}, as they are used in our experiments. These learned optimizers construct input features $\vu_t$ based on momentum accumulators, a variance accumulator, and multiple adafactor accumulators, we provide a full list in Tables~\ref{apdx:table:features-lopt},~\ref{apdx:table:features-velo-adafac}, and~\ref{apdx:table:features-velo-lstm} of the Appendix. At every gradient descent step, small\_fc\_lopt and VeLO are applied to each parameter of the optimizee, producing two outputs: the magnitude ($m$) and direction ($d$) of the update. VeLO additionally outputs a tensor-level learning rate, $\alpha_\mW$. The per-parameter update for both optimizers is given as 
\begin{equation}\label{eq:loupdate}
    w_t = w_{t-1} - \alpha_\mW \lambda_1 d \exp{\left( \lambda_2 m \right)},
\end{equation} 
where $w$ is a parameter of weight matrix $\mW$, $\lambda_1$ and $\lambda_2$ are constant values set to $0.001$ to bias initial step sizes to be small. 
For small\_fc\_lopt, $\alpha_\mW=1$ always. We refer interested readers to appendix sections~\ref{apdx:sec:smallfc} and~\ref{apdx:sec:velo} for more details.

\section{Related Work}
\label{sec:related-work}
\textbf{Generalization in LOs.} There are three main difficulties of learned optimizer generalization~\citep{chen2022learning,amos2022tutorial}: (1) optimizing unseen tasks; (2) optimizing beyond maximum unroll length seen during meta-training; (3) training optimizees that do not overfit.
Among these, (3) has been most extensively addressed as this problem has been well studied in classic optimization literature. For example, extra-regularization terms can be directly applied to a learned optimizer ~\citep{harrison2022closer,yang2023learning}. In addition, (3) can be addressed by meta-training on a validation set objective~\citep{metz2019understanding} or parameterizing LOs as hyperparameter controllers~\citep{almeida2021generalizable}. The problem (2) has been mitigated by regularization~\citep{harrison2022closer,yang2023learning} and larger-scale meta-training~\citep{velo}. However, (1) has remained a more difficult and understudied problem.

To the best of our knowledge, the only current approach to tackle this problem is to meta-train LOs on thousands of tasks~\citep{velo}. However, this approach is extremely expensive and seems bound to fail in the regime where the optimizer is expected to generalize from small meta-training tasks in standard parameterization to large unseen tasks: figures $6$ and $9$ of~\citet{velo} demonstrate that this was not achieved even when using $4000$ TPU-months of compute. Generalization would be expected if all tasks, no matter the size, were included in the meta-training distribution, but such an approach is simply intractable and is likely to remain so.

\textbf{Maximal Update Parametrization and Hyperparameter transfer.}
First proposed by~\citet{tensorprogramsiva}, the Maximal Update Parametrization is the unique stable abc-Parametrization where every layer learns features. The parameterization was derived for adaptive optimizers by~\citet{tesnorprogramsivb} and was applied by~\citet{tensorv} to enable zero-shot hyperparameter transfer for Adam and SGD. Most recently, in tensor programs VI,~\citet{mudepth} propose Depth-$\mu$P, a parameterization allowing for hyperparameter transfer in infinitely deep networks. While it is appealing, Depth-$\mu$P is only valid for residual networks with a block depth of $1$, so it does not apply most practical architectures (e.g., transformers, resnets, etc.). For these reasons, we do not study Depth-$\mu$P herein. Following from the original discovery of hyperparameter transfer in~\cite{tensorv}, a number of follow-up works have emerged that are not part of the tensor programs series. \cite{sparsemup} investigates transferring hyperparameters across different sparsity levels and widths. \cite{umup} investigates a combination of $\mu$P and unit scaling, which results in easier tuning and more stable low-precision training. \cite{everett2024exponents} investigate the alignment assumptions of~\cite{tensorv} and find that appropriate per-layer learning rate prescriptions can also enable hyperparameter transfer in standard, mean field, and NTK parameterizations. In their empirical investigation of scaling exponents across these parameterizations, the authors find that SP with layer-wise learning rates outperforms $\mu$P. While we study the impact of meta-learning optimizers in $\mu$P on meta-generalization herein, it is still an open question which parameterization is best for meta-learning optimizers. Finally, in concurrent work,~\cite{completep} propose CompleteP, a parameterization that can achieve transfer of optimal hyperparameters across depth and width.

\section{$\mu$-parametrization for Learned Optimizers}\label{sec:proof}
Parameterizing an optimizee neural network in $\mu$P requires special handling of the initialization variance, pre-activation multipliers, and optimizer update for each weight matrix $\mW \in \R^{n \times m}$ in the network. Specifically, these quantities will depend on the functional form of the optimizer and the dependence of $n$ ({\small\textsc{fan\_out}})  and $m$ ({\small\textsc{fan\_in}}) on width. We will refer to weight matrices in a network of width $h$ as hidden layers if $\Theta(n) = \Theta(m) = \Theta(h)$, as output layers if $\Theta(n) = \Theta(1) \land \Theta(m) = \Theta(h)$, and as input layers if $\Theta(n) = \Theta(h) \land \Theta(m) = \Theta(1)$. Here, $\Theta$ is standard asymptotic notation. Note that all biases and the weights of normalization layers are considered input layers and should be scaled as such. With this in mind, consider an arbitrary neural network\footnote{The $\mu$LO parameterization can be applied to any neural network architecture.} whose weight matrices are denoted $\mW_l$, where $l$ indexes the layers; the following modifications are then required to obtain $\mu$P for learned optimizers.

\vspace{-10pt}
\paragraph{Optimizee Initialization-$\mu$.} If $\mW_l$ belongs to a hidden or input layer, its weights should be initialized as $\mathcal{N}(0,\frac{1}{\textsc{fan\_in}})$. Output layers should have their weights initialized as $\mathcal{N}(0,1)$.
\vspace{-10pt}
\paragraph{Optimizee Multipliers-$\mu$.} Output layer pre-activations should be multiplied by $\frac{1}{\textsc{fan\_in}}$ during the forward pass.
\vspace{-10pt}
\paragraph{Optimizer Update Scaling-$\mu$.} The learned optimizer's update (eq.~\ref{eq:loupdate}) is re-scaled as follows:
\begin{equation}\label{eq:update}
w_t = \begin{cases}w_{t-1} - \frac{1}{\textsc{fan\_in}} \cdot \left(\alpha_{_{\mW_l}}\lambda_1 d \exp{\left( \lambda_2 m\right)}\right) & \text{$\mW_l$ is a hidden layer}\\
     w_{t-1} - \alpha_{_{\mW_l}}\lambda_1 d \exp{\left( \lambda_2 m \right)} & \text{otherwise.} 
    \end{cases}
\end{equation}
Where $w$ is a parameter of the weight matrix, $\mW_l$, and the dependence of $d$ and $m$ on $w_{t-1}$ is not made explicit for simplicity. For transfer to the largest width optimizees, it may also become necessary to re-scale numerical underflow constants ($\epsilon$) by $\frac{1}{\textsc{fan\_in}}$ as suggested by~\citep{everett2024exponents}. However, for the scales reported on by our experiments, we did not find this to be necessary.

We now prove that our parameterization satisfies the $\mu$P Desiderata (\citep{tensorv} Sec. J.2.1).

\begin{proposition}[$small\_fc\_lopt$ $\mu$P]
Assume that the Learned Optimizer $f_\phi$ has the form $small\_fc\_lopt$ is fed with features given in Appendix~\ref{apdx:sec:smallfc} and that during training the optimizee's parameters and input data become aligned, leading to Law of Large Numbers (LLN) scaling, then the update, initialization, and pre-activation multiplier above is sufficient to obtain a Maximal Update Parametrization.
\end{proposition}

\begin{proposition}[VeLO $\mu$P]
Assume that $\phi$ in Proposition 4.1 is generated using an LSTM with the input features described in Appendix~\ref{apdx:sec:velo} and that during training the optimizee's parameters and input data become aligned, leading to Law of Large Numbers (LLN) scaling, then the update, initialization, and pre-activation multiplier above is sufficient to obtain a Maximal Update Parametrization.
\end{proposition}
\vspace{-5pt}
\textit{Proof.} The proof is provided in Appendix~\ref{apdx:sec:proof}. $\qed$


\section{Empirical evaluation}
\label{sec:results}
We construct a suite of optimization tasks of varying width to evaluate the meta-generalization properties of our $\mu$LOs meta-trained on MLPs vs per-task tuned $\mu$Adam~\citep{tensorv}, per-task tuned SP AdamW~\citep{adamw}, and baseline SP LOs (meta-trained on MLP tasks).
Our main focus is to evaluate meta-generalization to wider networks as this is a key weakness of learned optimizers in previous works. However, we also establish the generalization properties of $\mu$LOs to deeper networks and longer training horizons. Please note that while $\mu$LOs inherit the theoretical properties of $\mu$P for width scaling, our findings with respect to deeper networks and longer training are purely empirical.

\subsection{Setup}

\noindent\textbf{Baseline LOs and $\mu$LOs.} The meta-training configuration of each learned optimizer is summarized in Table~\ref{tab:optimizers}. Each learned optimizer (ours and the baselines) in our empirical evaluation is meta-trained using the multiple-width single-task meta-training recipe proposed in section~\ref{sssec:metadist}. \textbf{Notably, these tasks only include MLPs (see Fig~\ref{fig:metagen}), while the hand-desinged optimizers in our study are tuned individually on each task.} The SP baselines sheds light on whether simply varying the SP optimizee width during meta-training is enough to achieve generalization of the LO to wider networks in SP. During meta-training, we set the inner problem length to be $1000$ iterations. Therefore, any optimization beyond this length is considered out-of-distribution. For all meta-training and hyperparameter tuning details, including ablation experiments, see section~\ref{apdx:sec:meta-train} of the appendix.

\noindent\textbf{$\mu$Adam} is a strong hand-designed $\mu$P baseline. It follows the Adam $\mu$-parametrization and does not use weight decay as this is incompatible with $\mu$P~\citep{tensorv}. $\mu$Adam is tuned on a width=$1024$ version of each task as this is the width of the largest meta-training task seen by our learned optimizers (see Table~\ref{tab:optimizers}). We tune the learning rate ($\eta$) and accumulator coefficients ($\beta_1$ and $\beta_2$) using a grid search over more than $500$ different configurations. This is repeated once for each task in our suite. Section~\ref{sec:apdx:muadam-tuning} of the appendix provides more details about the grid search including the values swept and the best values found.


\noindent\textbf{AdamW}~\citep{adamw} is a strong hand-designed SP baseline. It is tuned on the largest meta-training task seen by our learned optimizers (Table~\ref{tab:optimizers}).  AdamW is tuned on a width=$1024$ version of each task as this is the width of the largest meta-training task seen by our learned optimizers (see Table~\ref{tab:optimizers}). We tune the learning rate ($\eta$), accumulator coefficients ($\beta_1$ and $\beta_2$), and weight decay ($\lambda$) using a grid search over more than $500$ different configurations. This is repeated once for each task in our suite. Section~\ref{sec:apdx:adamw-tuning} of the appendix provides more details about the grid search including the values swept and the best values found.

\noindent\textbf{Evaluation tasks.} Our evaluation suite includes $35$ tasks spanning image classification (CIFAR-10, ImageNet) using MLPs and Vision Transformers (ViTs)~\citep{dosovitskiy2020image} and autoregressive language modeling with a decoder-only transformer on LM1B~\citep{chelba2013lm1b}. To create the tasks, we further vary image size (for image classification), width, and depth of the optimizee network, and the number of optimization steps. 
See Table~\ref{apdx:tab:testtasks} of the appendix for an extended description of all the tasks.

\begin{figure*}[t]
\includegraphics[width=0.85\linewidth]{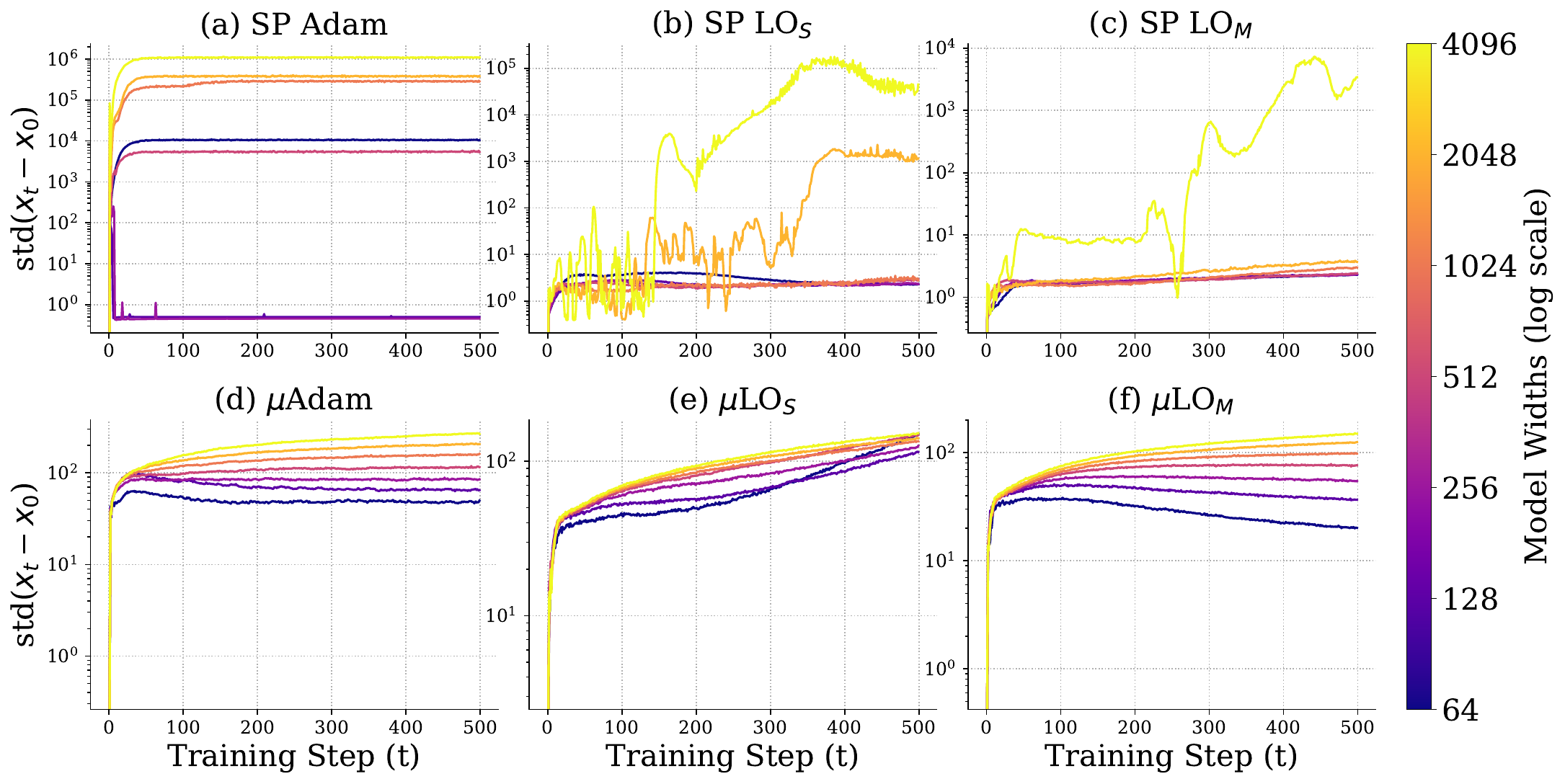}
    \centering
    \caption{\textbf{Layer 2 pre-activations behave harmoniously in \mup for $\mu$LOs and $\mu$Adam alike.} We report the evolution of coordinate-wise standard deviation of the difference between the initial ($t=0$) and $t$-th second-layer pre-activations of an MLP during training for the first $500$ steps of a single run (the remaining layers behave similarly, see Sec.~\ref{sec:apdx:activations}). We observe that all models parameterized in \mup enjoy stable coordinates across widths, while the pre-activations of larger-width models in SP blow up after a number of training steps. }
    \vspace{-10pt}
    \label{fig:layer2pa}
\end{figure*}

\vspace{-12pt}
\subsection{Results}
\vspace{-8pt}

In the following sections, we evaluate different meta-training distributions for training $\mu$LOs (Sec.~\ref{sssec:metadist}); we present results empirically verifying the pre-activation stability of our $\mu$LOs (Sec.~\ref{sssec:stability}); we present the results of our main empirical evaluation of meta-generalization to wider networks (Sec.~\ref{sssec:stability}); a study of $\mu$LOs generalization to deeper networks  (Sec.~\ref{sssec:beyond}); and a study of $\mu$LOs generalization to longer training horizons  (Sec.~\ref{sssec:beyond}). All of our figures report training loss and report the average loss across $5$ random seeds. Each seed corresponds to a different ordering of training data and a different initialization of the optimizee. All error bars in our plots report standard error across seeds. Standard error is $\frac{\sigma}{\sqrt{n}}$ where $\sigma$ is the population standard deviation and $n$ is the number of samples.

\vspace{-5pt}
\subsubsection{Evaluating meta-training distributions for $\mu$LOs} 
\begin{figure}
    \centering
    \subfloat[Iteration $1000$]{\includegraphics[width=0.25\linewidth]{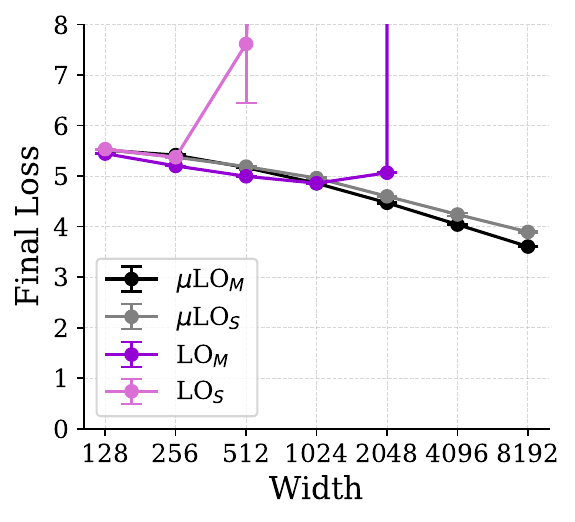}}
    \subfloat[Iteration $5000$]{\includegraphics[width=0.25\linewidth]{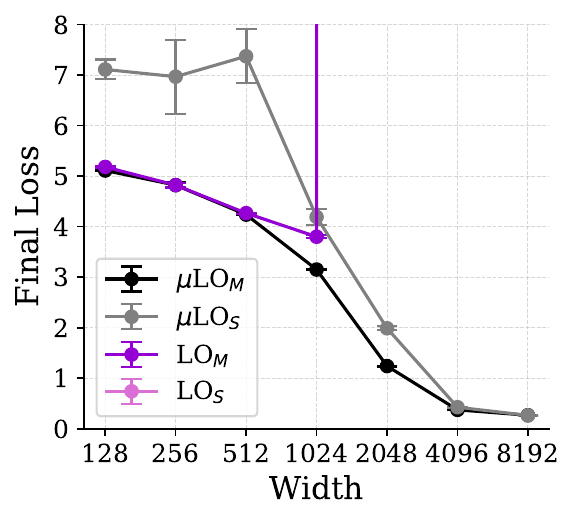}}
    \vline
    \subfloat[$64\times64\times3$ ImageNet]{\includegraphics[width=0.25\linewidth]{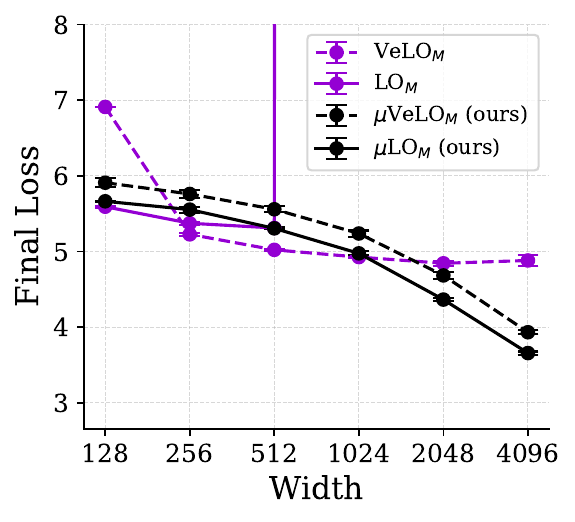}}
    \subfloat[$32\times32\times3$ Cifar10]{\includegraphics[width=0.25\linewidth]{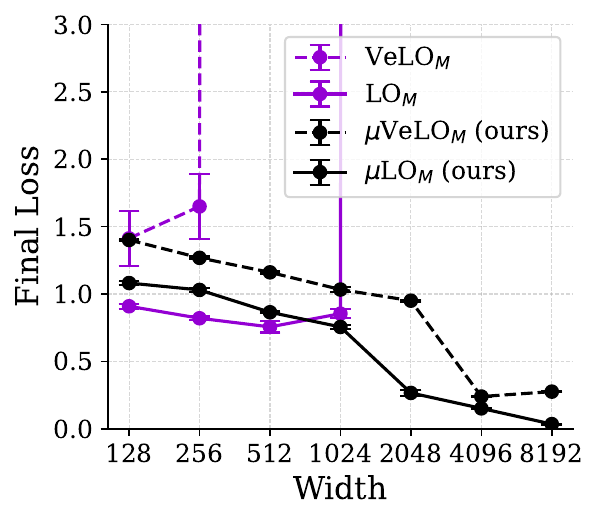}}
    \caption{\textbf{Generalization beyond meta-training widths is severely limited without our approach.}
    Each point is the average final training loss over $5$ seeds with standard error bars. Subfigures (a) and (b) report the results of our meta-training task ablation on the ImageNet-32 meta-training tasks at 1000 and 5000 steps. Subfigures (c) and (d) report the performance of $\mu$LO$_M$ and $\mu$VeLO$_M$ on OOD datasets.}
    
    \label{fig:task-number-ablation}
\end{figure}

\vspace{-5pt}
\label{sssec:metadist}
In $\mu$-transfer~\citep{tensorv}, hyperparameters are typically tuned on a small proxy task before being transferred to the large target task. In contrast, learned optimizers are typically meta-trained on a distribution of tasks. To verify the effectiveness of each approach for meta-training $\mu$LOs, we compare $\mulos$, meta-trained on a single width=$128$ MLP ImageNet classification task (see Tab.~\ref{tab:optimizers}), to $\mulom$, meta-trained on width $\in\{128,512,1024\}$ MLP ImageNet classification tasks. Each optimizer targets $1000$ step problems. We include equivalent standard parameterization baselines for reference ($\los$ and $\lom$). Figure~\ref{fig:task-number-ablation} reports the performance of each optimizer on a suite of MLP 
classification tasks of increasing width. When training for $1000$ steps (meta-training unroll length), we observe that $\mulom$ outperforms $\mulos$ as the width of the model is increased (Fig.~\ref{fig:task-number-ablation} (a)). Moreover, we observe that there is a discrepancy in performance between both models after $5000$ steps (Fig.~\ref{fig:task-number-ablation} (b)), showing that meta-training with multiple tasks of different widths has benefits for generalization to longer unrolls in addition to improved generalization to larger optimizees. Given the improved generalization of $\mulom$ compared to $\mulos$, we adopt the multiple-width meta-training recipe as part of our method. Subsequent experiments (e.g., Figures~\ref{fig:task-number-ablation} and~\ref{fig:results-width}) will show that our recipe is also effective for meta-training $\mu$VeLO.

\vspace{-5pt}
\subsubsection{Evaluating pre-activation stability}\label{sssec:stability}
We now verify that desiderata J.1 of \cite{tensorv} is satisfied empirically. In Figure~\ref{fig:layer2pa}, we report the evolution of the coordinate-wise standard deviation of the difference between initial (t=0) and current (t) second-layer pre-activations of an MLP during the first $500$ steps of training for a single trial. We observe that all models parameterized in \mup enjoy stable coordinates across widths, suggesting that desiderata J.1 is satisfied by our parameterization. In contrast, the pre-activations of the larger MLPs in SP blow up immediately for SP Adam while they take noticeably longer for LO$_S$ and LO$_M$. 
Section~\ref{sec:apdx:activations} of the appendix contains similar plots for the remaining layers of the MLP which show similar trends. In summary, we find, empirically, that pre-activations of $\mu$LOs and $\mu$Adam are similarly stable across widths, while the activations of SP Adam and SP LOs both blow up but behave qualitatively differently.

\vspace{-5pt}
\subsubsection{Meta-generalization to wider networks}
\begin{figure*}[t]
    \centering
    \subfloat[MLP IN32 W=$8192$ ]{\includegraphics[width=0.33\linewidth]{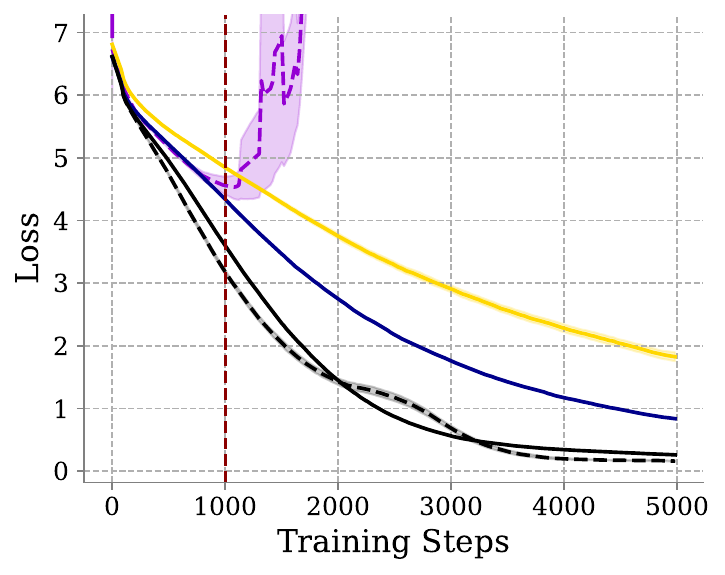}}
    \subfloat[MLP IN64 W=$8192$]{\includegraphics[width=0.33\linewidth]{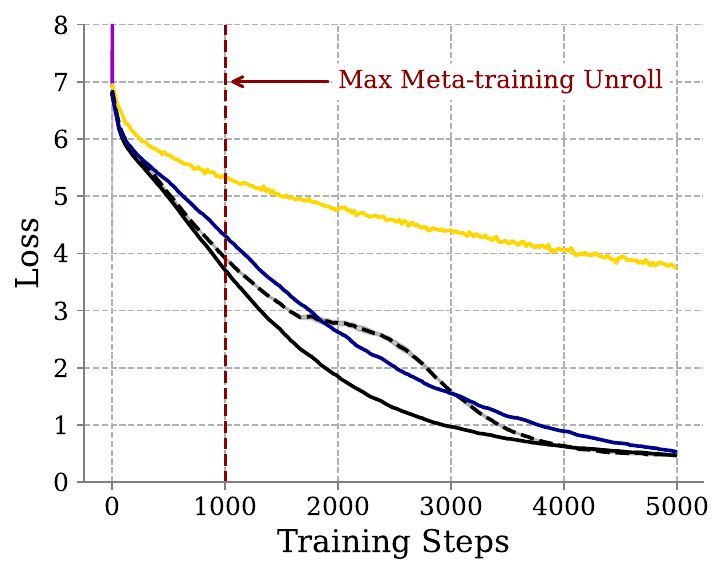}}
    \subfloat[MLP C10 W=$8192$ ]{\includegraphics[width=0.33\linewidth]{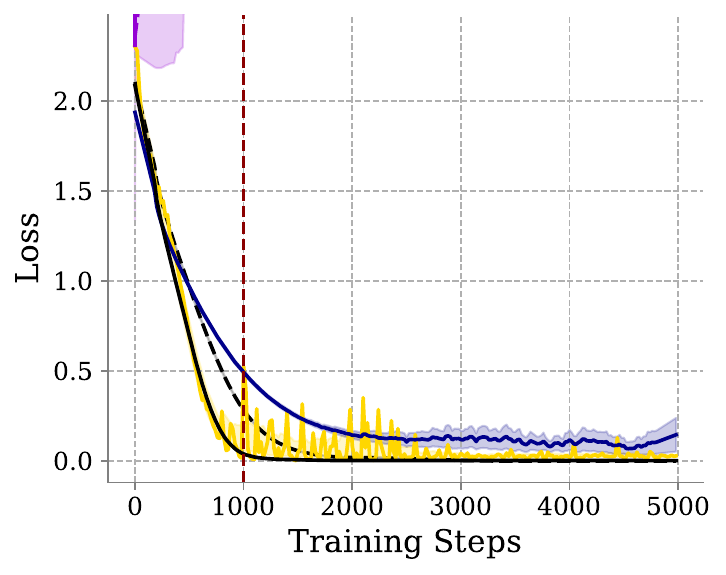}}\\
    \vspace{-5pt}
    \raisebox{0.6cm}{\includegraphics[width=0.22\linewidth]{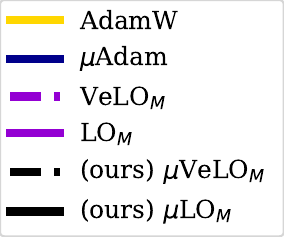}}
    \subfloat[LM W=$4096$]{\includegraphics[width=0.33\linewidth]{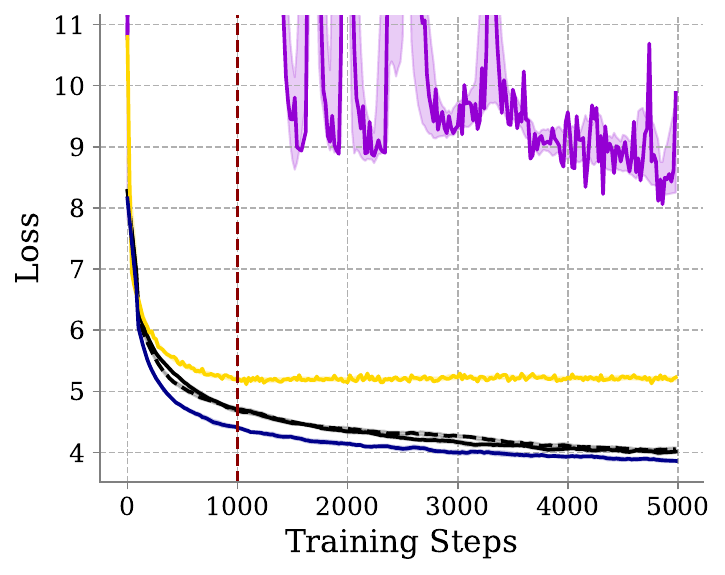}}
    \subfloat[ViT W=$4096$]{\includegraphics[width=0.33\linewidth]{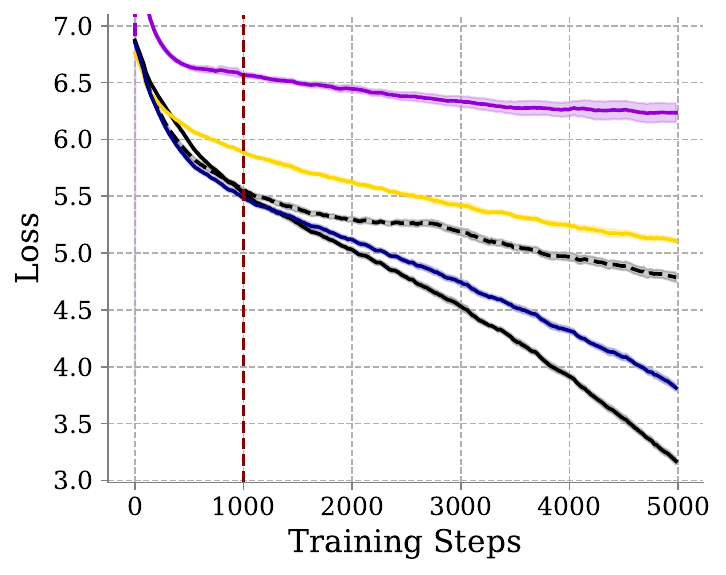}}

    \caption{\textbf{Evaluating generalization to wider networks for different tasks.} All optimizers are meta-trained or hyperparameter tuned for $1000$ inner steps (dotted red line), therefore, any optimization beyond $1000$ steps is considered out-of-distribution. We plot average training loss over $5$ seeds with standard error bars. We observe that $\mulom$ and $\mu$VeLO$_M$generalize smoothly to longer unrolls and all unseen tasks, unlike their SP counterparts which diverge or fail to make progress. $\mu$LOs outperform the extensively tuned AdamW and $\mu$Adam baselines in subfigures (a),(b), match or surpass them in subfigure (c), and exceed or nearly match their performance on far out-of-distribution LM and ViT tasks (subfigures (d) and (e)). Note that all AdamW and $\mu$Adam are tuned on smaller versions of each task, while our $\mu$LOs are only meta-trained on MLP tasks. \vspace{-10pt}
    }
    \label{fig:results-width}
\end{figure*}

\vspace{-4pt}
\label{sssec:wider-networks}
Given our goal of improving LO generalization to unseen wider tasks, the bulk of our empirical evaluation is presented in this section. Specifically, we evaluate the behavior of $\mu$LOs as the width of tasks increases well beyond what was seen during meta-training. To accomplish this, we fix the depth of each task and vary the width (see Table~\ref{apdx:tab:testtasks} for a full list of tasks), leading to a testbed of $32$ different tasks. We then train each task using the baselines and $\mu$-optimizers outlined in section~\ref{sec:results} for $5000$ steps for $5$ different random seeds. This involves training $1120$ different neural networks. To make the results easily digestible, we summarize them by width and final performance in Figure~\ref{fig:results-width} and by average optimizer rank in Table~\ref{tab:results-tab}. We also highlight the smooth training dynamics of our optimizers at the largest widths in Figure~\ref{fig:results-width}.
\vspace{-8pt}
\paragraph{Performance measured by final loss as a function of width.}Figure~\ref{fig:task-number-ablation} compares the training loss after $1000$ steps of SP learned optimizers to $\mu$-parameterized learned optimizers for different widths. This is shown in three subfigures for three MLP image classification tasks: (a) Imagenet $32\times 32\times3$ (IN32), (c) Imagenet $64\times 64\times3$ (IN64), and (d) Cifar-10 $32\times 32\times3$ (C10). Subfigure (a) shows the performance of learned optimizers on larger versions of the meta-training tasks. We observe that the $\mu$LOs achieve lower final training loss as the width of the task is increased. In contrast, $\lom$ diverges for widths larger than $2048$. 
Subfigure (b) evaluates our $\mu$LOs on $64\times 64\times3$ ImageNet images (e.g., when the input width is larger). Similarly, we observe smooth improvements in the loss as the optimizee width increases for $\mu$LOs, while their SP counterparts either diverge at width $512$ (LO$_M$) or fail to substantially improve the loss beyond width $1024$ (VeLO$_M$). Finally, Subfigure (c) shows the performance of our $\mu$LOs on Cifar-10 (smaller output width) as the width is increased. Similarly, we observe smooth improvements in the loss as the width increases for $\mu$LOs, while their SP counterparts either diverge immediately at small widths (VeLO$_M$) or diverge by width $1024$ (LO$_M$).

\paragraph{Training dynamics at the largest widths}Figure~\ref{fig:results-width} reports the training curves of different optimizers on the largest width tasks in our suite. Despite training for $5\times$ longer than the maximum meta-training unroll length, our $\mu$LOs are capable of smoothly decreasing the loss for the largest out-of-distribution tasks in our suite. In contrast, the strong SP LO baselines diverge by $1000$ steps (subfigures (a),(b),(c),(d)), or fail to decrease the training loss (subfigure (e)), demonstrating the clear benefit of $\mu$LOs for learned optimization. Our $\mu$LOs also substantially best the per-task-tuned AdamW and $\mu$Adam baselines (subfigures (a) and (b)), match the best performing hand-designed optimizer in subfigure (c), and nearly matches or outperforms the strongest hand-designed baseline performance on far out-of-distribution LM and ViT tasks (subfigures (d) and (e)). These results demonstrate that, under our $\mu$LO meta-training recipe, learning optimizers that smoothly train large neural networks (e.g., demonstrated an 8B parameter model typically uses width=$4096$) is possible at low cost ($\mulom$ is meta-trained for 100 GPU hours). 

\begin{table*}[h]
\centering
\caption{\textbf{Summary of optimizer performance on large tasks.} We report the average rank of different optimizers across the five tasks in our suite. We evaluate each optimizer on large-width tasks: Large (2048), XL (4096 for MLPs and 3072 for vit and LM), and XXL (largest size for each task see Tab.\ref{apdx:tab:testtasks} of the appendix). We bold the strongest, underline the second strongest, and italicize the third strongest average rank in each column. We observe that, across all iterations, $\mulom$ and $\muvelom$ consistently obtain the best and second-best ranks for all tasks. \vspace{-10pt}
}
\label{tab:results-tab}
\begin{adjustbox}{width=\columnwidth,center}
\begin{tabular}{l|ccc|ccc|ccc}
\toprule
& \multicolumn{3}{c|}{\textbf{Loss at 1k steps}} & \multicolumn{3}{c|}{\textbf{Loss at 3k steps}} & \multicolumn{3}{c}{\textbf{Loss at 5k steps}} \\
\textbf{Optimizer}     & \textbf{OoD (Large)} & \textbf{OoD (XL)}  &  \textbf{OoD (XXL)}&   \textbf{OoD (Large)}    & \textbf{OoD (XL)}      &  \textbf{OoD (XXL)}& \textbf{OoD (Large)} & \textbf{OoD (XL)}  &  \textbf{OoD (XXL)} \\ \midrule
AdamW                  & \textit{3.00}   & 3.60              & 4.40            & \textit{2.80}   & 2.60             & 4.00                   & \textit{2.60}  & \textit{2.40} & 3.80 \\
$\mu$Adam              & 3.40             & \textit{2.20}    & \textit{2.20}   & 3.00            & \textit{2.40}    & \textit{2.40}          & 3.20 & 2.60 & \textit{2.60} \\
VeLO$_M$               & 4.60             & 4.00             & 5.00            & 5.40            & 5.40             & 5.80                   & 6.00 & 5.40 & 5.80 \\
LO$_M$                 & 5.60             & 5.40             & 5.60            & 5.60            & 4.80             & 5.20                   & 5.00 & 4.80 & 5.20 \\
$\mu$VeLO$_M$ (ours)   & \underline{2.60} & \textbf{1.60}    & \textbf{1.80}   & \underline{2.40}& \underline{2.00} & \underline{2.40}      & \underline{2.40} & \textbf{1.40} & \underline{2.00} \\
$\mu$LO$_M$ (ours)     & \textbf{1.80}    & \underline{2.00} & \underline{2.00} & \textbf{1.80}  & \textbf{1.60}    & \textbf{1.20}         & \textbf{1.80} & \underline{2.20} & \textbf{1.60} \\
\bottomrule
\end{tabular}
\end{adjustbox}\vspace{-10pt}
\end{table*}
\vspace{-6pt}
\paragraph{Performance measured by average optimizer rank}Table~\ref{tab:results-tab} reports the average rank of different optimizers on out-of-distribution w.r.t. width tasks (Large (width 2048), XL (width $3072$ for transformer and $4096$ for MLPs), and XXL (maximum width)). Each entry of the table corresponds to the optimizer's average rank (within the $6$ optimizers evaluated) over the $5$ tasks in our suite: Cifar 10 MLP image classification, ImageNet 32 MLP image classification, ImageNet 64 MLP image classification, ImageNet 32 ViT image classification, and LM1B transformer language modeling. The optimizers are ranked by their training loss at the given iteration. We report average ranks for $1000$ iterations (inner-problem length), $3000$ iterations, and $5000$ iterations. We \textbf{bold} the strongest, \underline{underline} the second strongest, and \textit{italicize} the third strongest average rank in each column.  We observe that, across all iterations and all task sizes (Large, XL, XXL), either $\mulom$ or $\muvelom$ consistently obtain the best and second-best ranks for all tasks. The per-task-tune hand-designed baselines consistently occupy third and fourth rank, while the SP learned optimizer baselines perform worst, typically failing to optimize at this size. These results demonstrate that meta-training learned optimizers under the $\mu$-parameterization we propose and using our simple meta-training recipe yields substantial improvements in meta-generalization (across various tasks and widths) over SP LOs (previous work) and strong per-task tuned hand-designed baselines.



\begin{figure*}[h]
\vspace{-10pt}
    \centering
    \subfloat[ViT W=$1024$ D=$16$ ]{\includegraphics[width=0.28\linewidth]{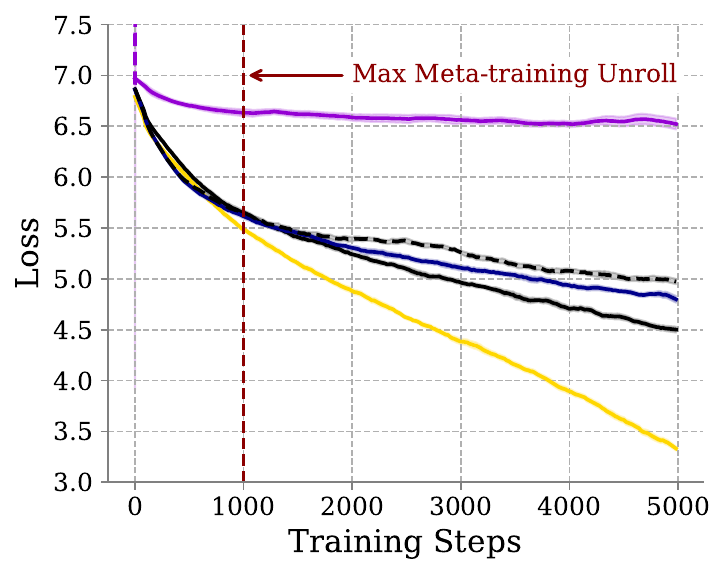}}
    \subfloat[LM W=$1024$ D=$16$ ]{\includegraphics[width=0.28\linewidth]{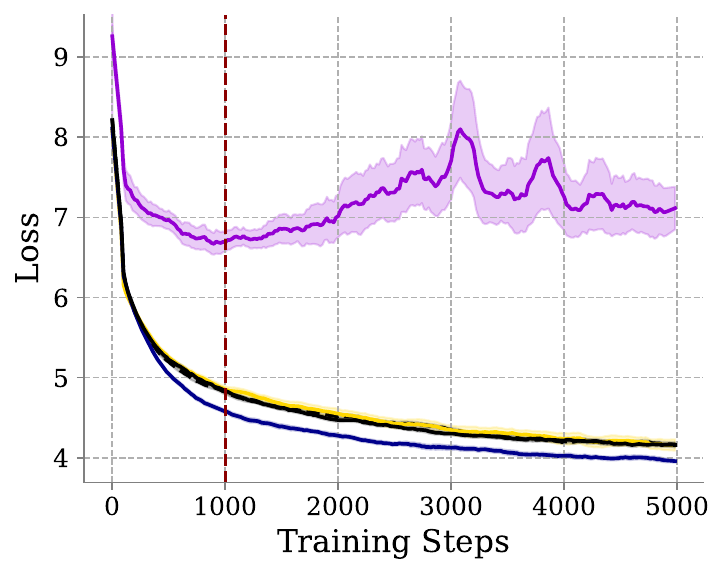}}
    \subfloat[MLP IN32 W=$1024$ D=$16$]{\includegraphics[width=0.28\linewidth]{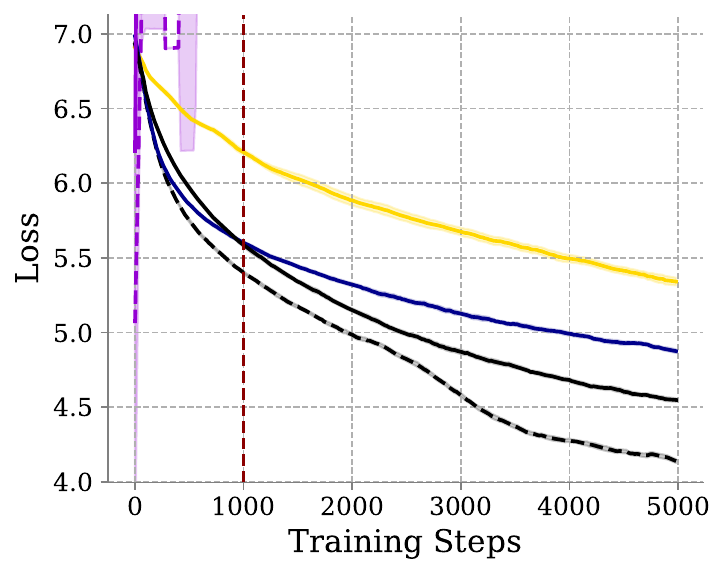}}
    \raisebox{0.7cm}{\includegraphics[width=0.16\linewidth]{figs/no_velo/legend_only_oracle.pdf}}\\
    \vspace{-5pt}
    \caption{\textbf{Evaluating generalization capabilities of $\mu$LOs to deeper networks}. Our focus is on comparing the meta-generalization to deeper tasks of $\mu$LOs to SP LOs (all meta-trained exclusively on MLPs). We also report the performance per-task tuned AdamW and $\mu$Adam for reference. Each plot reports average training loss over $5$ seeds with standard error bars. In each case, $\mu$LOs show improved generalization and performance when compared to their SP counterparts. 
    }
    \label{fig:results-depth}
\end{figure*}
\vspace{-4pt}
\subsubsection{Evaluating Meta-generalization Beyond Width}
\label{sssec:beyond}
\vspace{-4pt}
While our main focus is meta-generalization to wider networks
While the focus of our paper is improving the meta-generalization of LOs on wider tasks, it is also important to evaluate how these modifications to learned optimizer meta-training impact other axes of generalization. As such, we now study meta-generalization to deeper networks and longer training. While we provide strong AdamW and $\mu$Adam baselines for reference, our focus will be to establish the relative performance $\mu$LOs to SP LOs. Note that $\mu$P theory leveraged by $\mu$LOs specifically concerns transferring hyperparameters to larger-width networks, not longer training horizons or deeper networks. Therefore, any improvements we observe are purely empirical.

\textbf{Meta-generalization to deeper networks} In this section, we evaluate LO meta-generalization to deeper networks. Specifically, we increase the number of layers used in MLP, ViT, and LM tasks from $3$ to $16$, while keeping width=$1024$ within the range of tuning/meta-training. Figure~\ref{fig:results-depth} reports the performance of our learned optimizers on deeper networks. We observe that both $\mulom$ and $\muvelom$ optimize stably throughout and generally outperform their counterparts, $\lom$ and $\velom$, by the end of training on each task, despite being meta-trained on MLPs of exactly the same depth. Moreover, $\lom$ immediately diverges when optimizing the deep MLP while $\mulom$ experiences no instability. Similarly, $\velom$ diverges on ViTs and Transformers, while $\muvelom$ performs well, especially on ViTs. This is remarkable as, unlike width, there is no theoretical justification for \mup's benefit to deeper networks. We hypothesize that \mup's stabilizing effect on the optimizee's activations leads to this improvement in generalization (see Sec.~\ref{apdx:sec:explain} for more details).

\begin{figure*}[t]
    \centering
    
    \subfloat[ViT W=$1024$]{\includegraphics[width=0.28\linewidth]{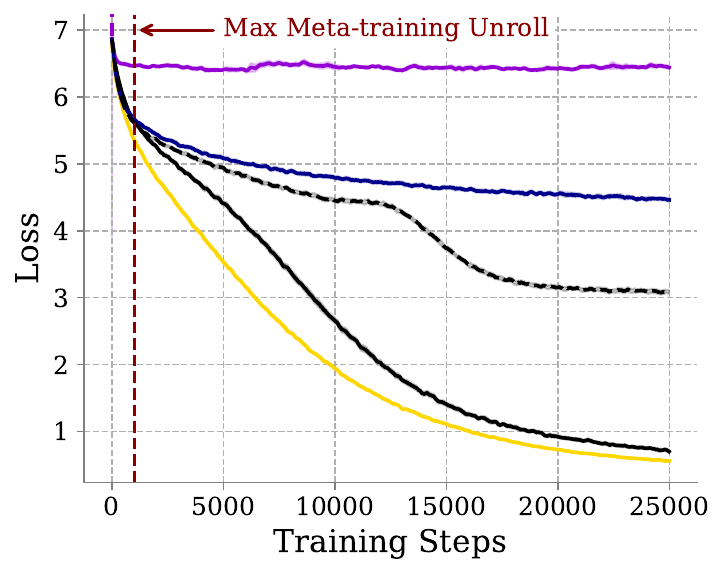}}
    \subfloat[LM W=$1024$]{\includegraphics[width=0.28\linewidth]{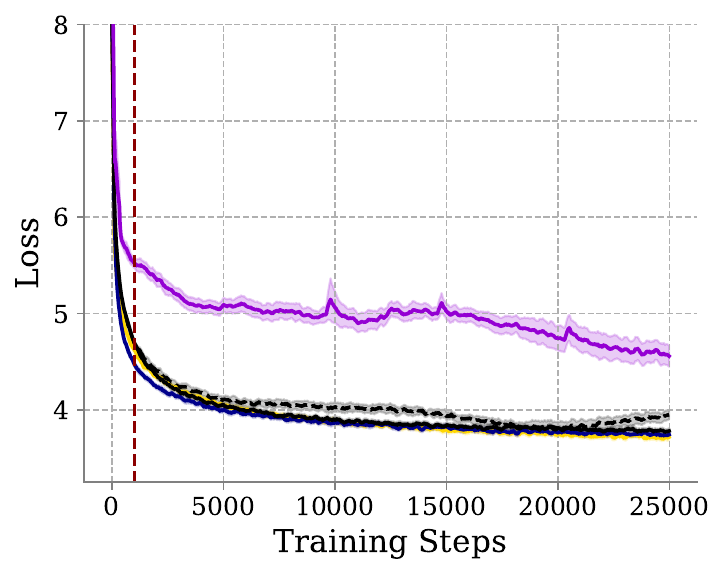}}
    \subfloat[MLP IN32 W=$1024$]{\includegraphics[width=0.28\linewidth]{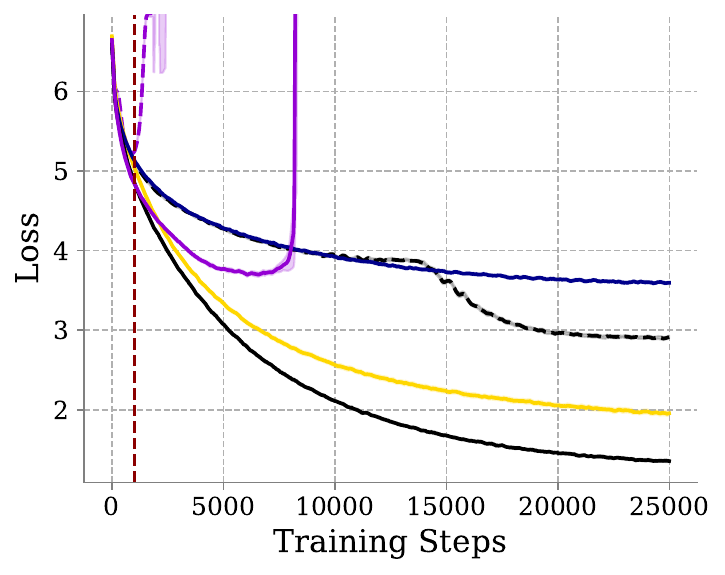}}
    \raisebox{0.7cm}{\includegraphics[width=0.16\linewidth]{figs/no_velo/legend_only_oracle.pdf}}\\
    \vspace{-5pt}
    \caption{\textbf{Evaluating meta-generalization to longer training horizons}.
    Note that AdamW and $\mu$Adam are evaluated on their tuning tasks here, while LOs are trained on MLPs. We plot average training loss over $5$ seeds with standard error bars. We observe that $\mu$LOs seamlessly generalize to training horizons 25$\times$ longer than meta-training. In contrast, the best performing SP LO fails to decrease training loss (a), decreases it but suffers instabilities (b), or diverges after $8000$ steps (c).
    \vspace{-15pt}
    }
    \label{fig:results-25k}
\end{figure*}

\textbf{Meta-generalization to longer training} In this subsection, we empirically evaluate the capability of $\mu$LOs to generalize to much longer training horizons than those seen during meta-training. Specifically, we use $\mulom$ and $\lom$ as well as $\muvelom$ and $\velom$ to train three networks with width $w=1024$: a 3-layer MLP, ViT on $32\times 32\times 3$ ImageNet and a 3-layer Transformer for autoregressive language modeling on LM1B. Each model is trained for $25,000$ steps (25$\times$ the longest unroll seen at meta-training time). Figure~\ref{fig:results-25k} reports the training loss averaged over $5$ random seeds. We observe that $\mulom$ and $\muvelom$ stably decrease training loss over time for each task, while $\lom$ and $\velom$ fail to decrease training loss (a), decreases it but becomes unstable (b), or diverges after $8000$ steps (c). While we are uncertain of the exact cause of this improved generalization, we hypothesize that it may be due to the improved pre-activation stability (see Sec.~\ref{apdx:sec:explain} for more details). These results suggest that generalization to longer training horizons is another benefit of using $\mu$LOs.


\vspace{-8pt}
\section{Limitations} 
\vspace{-4pt}
We have conducted a systematic empirical study and shown strong results within the scope of our study, there are some limitations of our work. Specifically, (1) we do not meta-train on tasks other than MLPs for image classification, (2) we do not provide an evaluation of models wider than $8192$ (MLPs) and $3072$/$12288$ (transformer hidden/FFN size) due to computational constraints in our academic environment, and (3) We did not include an oracle SP AdamW baseline whose hyperparameters are swept at every width due to computational constraints in our academic environment. 

\vspace{-4pt}
\section{Conclusion} %
\vspace{-4pt}
We have theoretically and empirically demonstrated that it is possible to obtain a valid $\mu$-parameterization for two state-of-the-art learned optimizer architectures. Under or proposed meta-training recipe, meta-learned optimizers show substantial improvements in meta-generalization properties when compared to strong baselines from previous work. Remarkably, our $\mu$LOs, meta-trained only on MLP tasks, surpass the performance of per-task-tuned hand-designed baselines in terms of average rank on wide OOD tasks. Moreover, our experiments also show that $\mu$LOs meta-trained with our recipe generalize better to wider and, unexpectedly, deeper out-of-distribution tasks than their SP counterparts. When evaluated on much longer training tasks, we observe that $\mu$LOs have a stabilizing effect, enabling meta-generalization to much longer unrolls ($25\times$ maximum meta-training unroll length). All of the aforementioned benefits of $\mu$LOs come at \emph{zero} extra computational cost compared to SP LOs. Our results outline a promising path forward for low-cost meta-training of learned optimizers that can generalize to large unseen tasks.

In future work, it will be important to investigate the benefits of meta-learning optimizers under parameterizations other than $\mu$P that have been shown to admit hyperparameter transfer~\citep{everett2024exponents}. Another important direction of inquiry is to investigate the meta-learning optimizers under parameterizations, like CompleteP~\citep{completep}, that have the potential to improve meta-generalization across depth and width. Finally, combining such parameterizations with improved meta-generalization and scalable meta-learning recipes is required for learning truly general-purpose optimizers.


\subsubsection*{Acknowledgments}
We acknowledge support from the Mila-Samsung Research Grant, FRQNT New Scholar [\emph{E.B.}], the FRQNT Doctoral (B2X) scholarship [\emph{B.T.}], the Canada CIFAR AI Chair Program [\emph{I.R.}], and the Canada Excellence Research Chairs Program in Autonomous AI [\emph{I.R.}]. We also acknowledge resources provided by Compute Canada, Calcul Québec, and Mila.  [\emph{E.O.}] acknowledges funding from PEPR IA (grant SHARP
ANR-23-PEIA-0008). He was granted access to the
AI resources of IDRIS under the allocation 2025-
AD011015884R1.

\bibliography{ref}
\bibliographystyle{iclr2026_conference}

\appendix

\newpage
\appendix\clearpage
\onecolumn
\section{Proof of Proposition 4.1}\label{apdx:sec:a}

For the reader's convenience, we will first review the input, output, update, and scaling of the per-parameter $\texttt{small\_fc\_lopt}$~\cite{metz2022practical} learned optimizer as it is necessary background for understanding our proof. This corresponds to the architecture of the $\mu$LO$_M$, $\mu$LO$_S$, LO$_M$, and LO$_S$ optimizers used throughout our experiments. In section~\ref{apdx:sec:velo}, we will also review the input, output, update, and scaling of VeLO, the architecture used for $\muvelom$ and $\velom$. Note that the VeLO~\cite{velo} architecture uses an almost-identical $\texttt{small\_fc\_lopt}$ network to produce per-parameter updates. The main difference is that VeLO uses an LSTM to generate the parameters of $\texttt{small\_fc\_lopt}$ for each tensor in the network at each optimization step. 


\subsection{$\mulom$ and $\muvelom$ input, output, update, and scaling.}
\subsubsection{The $\texttt{small\_fc\_lopt}$ architecture}\label{apdx:sec:smallfc}

$\texttt{small\_fc\_lopt}$ maintains three different per-parameter momentum accumulators ($\mM_{t, i}$) and one variance accumulator ($\mV_t$). In addition, it also maintains six adafactor-style accumulators of the column-wise ($\vc_{t, i}$) and row-wise ($\vr_{t, i}$) mean of the squared gradient. The accumulator update is given as follows:
\begin{align}   
\mM_{t, i} =\;& \beta_i \mM_{t-1, i} + (1 - \beta_i) \nabla_t & i \in \{1,2,3\} \nonumber, \\
\mV_t =\;& \beta_4 \mV_{t-1} + (1 - \beta_4) \nabla_t^2, &\nonumber\\
\vr_{t, i} =\;& \beta_i \vr_{t-1, i} + (1 - \beta_i)\;\texttt{row\_mean}(\nabla_t^2), & i \in \{5,6,7\}, \nonumber\\
\vc_{t, i} =\;& \beta_i \vc_{t-1, i} + (1 - \beta_i)\;\texttt{col\_mean}(\nabla_t^2), & i \in \{5,6,7\}, \nonumber\\
\mU_t :=\;& [\mM_{t,1}, \mM_{t,2}, \mM_{t,3}, \mV_t, \vr_{t,5}, \vr_{t,6}, \vr_{t,7}, \vc_{t,5}, \vc_{t,6}, \vc_{t,7}]. \nonumber
\end{align}
Here, we slightly abuse notation and define $\mU_t$ to be the entire accumulator state for all parameters in the optimizee (column-wise and row-wise features are repeated for notational convenience). After updating these accumulators, $\texttt{small\_fc\_lopt}$ computes additional learned optimizer input features:
\begin{align}
\displaystyle \mF^{(\nabla)}_i = \;&\nabla_t \odot \sqrt{\frac{\frac{1}{m}\sum_{h=1}^m(\vr_{t, i})_h}{ \vr_{t, i} \vc_{t, i}^T}},\nonumber \\   
\mF^{(\mM)}_i=\;&\displaystyle\mM_{t,j} \odot \sqrt{\frac{\frac{1}{m}\sum_{h=1}^m(\vr_{t, i})_h}{ \vr_{t, i} \vc_{t, i}^T}}, \nonumber \\
\mR_t=\;&\left[\frac{1}{\sqrt{\vr_{t,5}}}, \frac{1}{\sqrt{\vr_{t,6}}}, \frac{1}{\sqrt{\vr_{t,7}}}, \frac{1}{\sqrt{\vc_{t,5}}}, \frac{1}{\sqrt{\vc_{t,6}}}, \frac{1}{\sqrt{\vc_{t,7}}},\frac{\mM_{t,1}}{\sqrt{v}}, \frac{\mM_{t,2}}{\sqrt{v}}, \frac{\mM_{t,3}}{\sqrt{v}}, \frac{1}{\sqrt{v}}\right],\nonumber\\
\mH_t =\;&\left[\mF^{(\nabla)}_1,\mF^{(\nabla)}_2,\mF^{(\nabla)}_3,\mF^{(\mM)}_1,\mF^{(\mM)}_2,\mF^{(\mM)}_3\right],\nonumber\\
\hat{\mA}_t =\;& \vtheta_t \odot \nabla_t \odot \mH_t \odot \mR_t \odot \mU_t. \nonumber
\end{align}
Where $\odot$ denotes matrix concatenation across the feature dimension, $\vtheta_t$ are the optimizee's parameters, $\nabla_t$ is the optimizee's gradient, $\mH_t$ are adafactor normalized features, and $\mR_t$ are reciprocal features. Note that $\hat{\mA}_t \in \R^{|\vtheta|\times 28}$. The features within a parameter tensor are now normalized by their RMS-norm. Let $\mW^{(j)}\in\R^{m\times n}$ be the optimizee's j'th tensor and take $\hat{\mA}^{(j)} \in \R^{mn\times 28}$ to be the features of this tensor at  timestep $t$. Each feature $i$ within $\hat{\mA}^{(j)}$ is then then normalized as follows:
\begin{align}
    \bar{\mA}_{:,i}^{(j)} = \frac{  \hat{\mA}_{:,i}^{(j)} }{ \sqrt{\frac{1}{mn}\sum_{h=1}^{mn} (\hat{\mA}_{h,i}^{(j)})^2 }}.
\end{align}
Finally, the normalized features $\bar{\mA}$ are concatenated with timestep embeddings from step $t$ to form the complete input features for \texttt{small\_fc\_lopt}:
\begin{align}
    \mT_t =\;& [\tanh{\left( \frac{t}{x} \right)} \;\text{for} \;\;x \in \{1, 3, 10, 30, 100, 300, 1000, 3000, 10000, 30000, 100000\}],\nonumber\\
    \mA_t =\;& \bar{\mA_t} \odot \mT_t.\nonumber
\end{align}
 Concretely, in~\cite{metz2022practical}, $\texttt{small\_fc\_lopt}$'s architecture is a two-hidden-layer $4$ hidden-dimension MLP with ReLU activations: $f_\vphi(\mA)=\mW_2(ReLU(\mW_1ReLU(\mW_0\mA+\vb_0)+\vb_1)+\vb_2$. At each step, the learned optimizer maps the input features for each parameter, $p$, in the optimizee to a two-dimensional vector, $[d,m]$. At step $t$, the learned optimizer update for all parameters $p$ is given as follows:
\begin{align}
f_\vphi(\mA_p) =& \;[d_{p}, m_{p}];\nonumber\\
p_t =& \;p_{t-1} - \lambda_1 d_{p} e^{\left( \lambda_2 m_{p} \right)}.
\end{align}

\newcommand{\clip}{\operatorname{clip}}
\newcommand{\RMS}{\operatorname{RMS}}

Where $\lambda_1=\lambda_2=0.001$ to bias initial steps towards being small. We will now show that the inputs to $\texttt{small\_fc\_lopt}$ scales like $\Theta(1)$ as $n\to\infty$. Let's first see that any RMS-normalized quantity (e.g., the input to \texttt{small\_fc\_lopt}) is $\Theta(1)$, which we will subsequently use in our proof of propositions 4.1 and 4.2.
\begin{definition}
Let $\mW\in\R^{m\times n}$ be the weight matrix of a neural network. 
Let $\vv\in\R^{mn}$ be a vector, whose entries are statistics of parameters in $\mW$. We call 
\begin{align}
\bar{\vv} =  \frac{\vv}{ \text{RMS}(\vv)}\;\;;\;\;\text{RMS}(\vv)= \sqrt{\frac{1}{mn}}\|\vv\|_2.
\end{align}
The RMS-normalized~\cite{rmsnorm} version of $\vv$.
\end{definition}

\begin{proposition}\label{prop:rms}
    Let $\vv\in\R^{mn}$ be a vector whose entries scale like $\Theta(f(n))$, where $f:\R \rightarrow \R$ is a continuous function. Then, the entries of the RMS-normalized counterpart of $\vv$, $\bar{\vv}\in\R^{mn}$ will scale like $\Theta(1)$.
\end{proposition}
\begin{proof}
    Let $\vv\in\R^{mn}$ be a vector and $\bar{\vv}\in\R^{mn}$ denote its RMS-normalized counterpart. Then,
    \begin{align}\label{eq:rms}
    \bar{\vv} =\frac{\vv}{\sqrt{\frac{1}{mn}\sum_{h=1}^{mn} {\vv}_{h}^2 }}
    \end{align}
    where the division is elementwise. From the definition of $\Theta$, we know there exist constants $c_1,c_2>0$ and $N\in\mathbb{N}$ such that for all $n\ge N$ and every $h\in\{1,\dots,mn\}$,
\[
c_1\,|f(n)|
\;\le\;|v_h|\;\le\;c_2\,|f(n)|.
\]
Thus we have:
\begin{align}
v_h^2 &\;\in\; \bigl[c_1^2\,f(n)^2,\;c_2^2\,f(n)^2\bigr],\nonumber\\
\sum_{h=1}^{mn}v_h^2&\;\in\;\bigl[mn\,c_1^2\,f(n)^2,\;mn\,c_2^2\,f(n)^2\bigr].\nonumber\\
\frac{1}{mn}\sum_{h=1}^{mn}v_h^2&\;\in\;\bigl[c_1^2\,f(n)^2,\;c_2^2\,f(n)^2\bigr],\nonumber\\
\sqrt{\frac{1}{mn}\sum_{h=1}^{mn}v_h^2}&\;\in\;\bigl[c_1\,|f(n)|,\;c_2\,|f(n)|\bigr]\;=\;\Theta\bigl(f(n)\bigr).\nonumber\\
\end{align}
Since both numerator and denominator of~\ref{eq:rms} are $\Theta(f(n))$, their ratio is $\bar{v}_h = \Theta(1) \text{ for each }h.$ This completes the proof.
\end{proof}

\begin{corollary}\label{coro:smallfc}
    Assuming that time features are independent of width $n$, the coordinates of the input features to $\texttt{small\_fc\_lopt}$, as defined above, are $\Theta(1)$ as $n\to\infty$.
\end{corollary}
\begin{proof}
    This follows directly from proposition~\ref{prop:rms} since all non-time features in $\texttt{small\_fc\_lopt}$ are RMS-normalized.
\end{proof}

\subsubsection{The $\texttt{VeLO}$ architecture }\label{apdx:sec:velo}
\textbf{VeLO} uses an LSTM hypernetwork to produce the parameters, $\vphi_\mW$, of a $\texttt{small\_fc\_lopt}$ optimizer for each weight matrix $\mW$ in the optmizee network. Therefore, VeLO has the same accumulators as $\texttt{small\_fc\_lopt}$. VeLO's LSTM also outputs a learning rate multiplier, $\alpha_\mW$. For a parameter $p$ of $\mW$, the update becomes:
\begin{align}
f_{\vphi_\mW}(\mA^*_p) =& \;[d_{p}, m_{p}];\nonumber\\
p_t =& \;p_{t-1} - \alpha_\mW\lambda_1 d_{p} e^{\left( \lambda_2 m_{p} \right)}.
\end{align}
Where $\mA^*_p$ is a slightly modified version of the features outlined in the previous section (see Tab.~\ref{apdx:table:features-velo-adafac} for details), crucially, the features $\mA^*_p$ are all RMS-normalized as illustrated in the previous section.

To produce $\phi_\mW$ and $\alpha_\mW$, VeLO's LSTM takes as input $9$ remaining time features ($\mT$), $9$ EMA loss features ($\mL$), a one-hot vector representing the tensor's rank, three momentum features ($\texttt{var\_mom}_k \text{ for } k \in \{1,2,3\}$), and two variance features (\texttt{mean\_rms} \texttt{var\_rms}). For our goal of understanding valid parameterizations for VeLO, the most important LSTM features are the variance and momentum features as they are the only features that require further analysis of width scaling:
\begin{align}
\hat{m}_k =& \frac{1}{mn}\sum_i^m\sum_j^n\frac{\mM^{(k)}_{i,j}}{\text{RMS}(\mW)}, \nonumber\\
\texttt{var\_mom}_k =&\, c_1\,\clip\!\Bigl(\log\Bigl[\,\tfrac{c_2}{mn}\sum_{i,j}\Bigl(\tfrac{\mM^{(k)}_{ij}}{\RMS(\mW)}-\hat m_k\bigr)^2\Bigr],-\tau,\tau\Bigr),\nonumber \\
\texttt{mean\_rms} =&\, c_1\,\clip\!\Bigl(\log\Bigl[\,\tfrac{c_2}{mn}\sum_{i,j}\tfrac{\mV_{ij}}{\RMS(\mW)}\Bigr],-\tau,\tau\Bigr), \text{ and }\nonumber \\
\texttt{var\_rms}_k =&\, c_1\,\clip\!\Bigl(\log\Bigl[\,\tfrac{c_2}{mn}\sum_{i,j}\Bigl(\tfrac{\mV_{ij}}{\RMS(\mW)}-\hat m_k\Bigr)^2\Bigr],-\tau,\tau\Bigr).\nonumber
\end{align}
Where we set $c_1=\tfrac12, c_2=10,$ and $\tau=5$ following~\cite{velo}. Note that, in general, the quantities calculated within the log may not be nicely bounded, but since these features are clipped, straightforward analysis shows these features are $\Theta(1)$.

\begin{proposition}\label{prop:velo}
Let \(\mW\in\R^{m\times n}\) be a weight matrix whose entries scale as \(\Theta(n^p)\).  Let \(\hat m_k\), \(\texttt{var\_mom}_k\), \(\texttt{mean\_rms}\), and \(\texttt{var\_rms}_k\) be defined as above.
Assume \(\mM^{(k)}\) has the same per‐entry scaling as \(\mW\), and \(\mV\) has entries scaling as \(\Theta(n^{2p})\).  Then each of \(\hat m_k\), \(\texttt{var\_mom}_k\), \(\texttt{mean\_rms}\), and \(\texttt{var\_rms}_k\) is \(\Theta(1)\) as \(n\to\infty\).
\end{proposition}

\begin{proof}
First, observe that
\[
\RMS(\mW)
=\sqrt{\frac1{mn}\sum_{i,j}\mW_{i,j}^2}
=\sqrt{\Theta(n^{2p})}
=\Theta(n^p).
\]
Since \(\mM^{(k)}_{i,j}=\Theta(n^p)\), it follows that
\[
\frac{\mM^{(k)}_{i,j}}{\RMS(\mW)}
=\Theta\!\bigl(n^p/n^p\bigr)
=\Theta(1).
\]
Hence
\[
\hat m_k
=\frac1{mn}\sum_{i,j}\Theta(1)
=\Theta(1).
\]

Next, consider the argument of the logarithm in \(\texttt{var\_mom}_k\):
\[
\frac{c_2}{mn}
\sum_{i,j}\Bigl(\tfrac{\mM^{(k)}_{i,j}}{\RMS(\mW)}-\hat m_k\Bigr)^2.
\]
Each term \(\tfrac{\mM^{(k)}_{i,j}}{\RMS(\mW)}-\hat m_k\) is the difference of two \(\Theta(1)\) quantities, hence \(\Theta(1)\).  Summing \(mn\) such terms and dividing by \(mn\) yields \(\Theta(1)\).  Thus
\[
\log\Bigl[\tfrac{c_2}{mn}\sum_{i,j}\Bigl(\tfrac{\mM^{(k)}_{i,j}}{\RMS(\mW)}-\hat m_k\Bigr)\Bigr]
=\Theta(1),
\]
and clipping to \([-\tau,\tau]\) gives \(\Theta(1)\).  Multiplying by the constant \(c_1\) preserves \(\Theta(1)\).  Therefore \(\texttt{var\_mom}_k=\Theta(1)\).

For \(\texttt{mean\_rms}\), note \(V_{i,j}=\Theta(n^{2p})\), so
\[
\frac{\mV_{i,j}}{\RMS(\mW)}
=\Theta\!\bigl(n^{2p}/n^p\bigr)
=\Theta(n^p).
\]
Hence
\[
\frac{c_2}{mn}\sum_{i,j}\frac{\mV_{i,j}}{\RMS(\mW)}
=\Theta(n^p),
\]
and
\[
\log\bigl[\Theta(n^p)\bigr]
=\Theta(\log n).
\]
Clipping \(\log(n^p)\) to \([-\tau,\tau]\) yields a bounded constant \(\Theta(1)\), and multiplication by \(c_1\) gives \(\texttt{mean\_rms}=\Theta(1)\).

Finally, for \(\texttt{var\_rms}_k\), we have
\[
\frac{\mV_{i,j}}{\RMS(\mW)}-\hat m_k
=\Theta(n^p)-\Theta(1)
=\Theta(n^p),
\]
so
\[
\Bigl(\tfrac{\mV_{i,j}}{\RMS(\mW)}-\hat m_k\Bigr)^2
=\Theta(n^{2p}).
\]
Summing over \(mn\) entries and dividing by \(mn\) yields \(\Theta(n^{2p})\).  Taking the logarithm gives \(\Theta(\log n)\), clipping to \([-\tau,\tau]\) yields \(\Theta(1)\), and multiplying by \(c_1\) preserves \(\Theta(1)\).  Hence \(\texttt{var\_rms}_k=\Theta(1)\), completing the proof.
\end{proof}

\begin{corollary}\label{coro:velo}
    Assuming that time features are independent of width $n$, the coordinates of the input features to $\texttt{VeLO}'s$ LSTM, as defined above, are $\Theta(1)$ as $n\to\infty$.
\end{corollary}
\begin{proof}
    This follows directly from proposition~\ref{prop:velo} since all the other input features in $\texttt{VeLO}$ trivially $\Theta(1)$ as $n\to\infty$.
\end{proof}

\subsection{Proof: $\mu$-parameterization for learned optimizers}\label{apdx:sec:proof}



For the reader's convenience, we will now restate the $\mu$P desiderata (Appendix J.2~\cite{tensorv}) which will be used by our proof. When using a maximal update parameterization, at any point during training, the following conditions should be met:
\begin{enumerate}
    \item \textbf{(Activation Scale)} Every (pre-)activation vector $x\in\R^n$ in the network should have $\Theta(1)$-sized coordinates.
    \item \textbf{(Output Scale)} The output of the neural network $f_\theta(x)$ should be $O(1)$.
    \item \textbf{(Maximal Updates)} All parameters should be updated as much as possible without divergence. In particular, updates should scale in width so that each parameter has nontrivial dynamics in the infinite-width limit.
\end{enumerate}

 While we do not go into the level of mathematical detail of~\cite{tesnorprogramsivb}, our intention in propositions 4.1 and 4.2 is to show that the above desiderata are satisfied in practice by the two popular learned optimizer architectures we study.



\textbf{Proposition 4.1.} \textit{Assume that the Learned Optimizer $f_\phi$ has the form $small\_fc\_lopt$ is fed with features given in Appendix~\ref{apdx:sec:smallfc} and that during training the optimizee's parameters and input data become aligned leading to Law of Large Numbers (LLN) scaling, then  the update, initialization, and pre-activation multiplier above is sufficient to obtain a Maximal Update Parametrization.}

\textbf{Proposition 4.2.} \textit{Assume that $\phi$ in Proposition 4.1 is generated using an LSTM with the input features described in Appendix~\ref{apdx:sec:velo} and that during training the optimizee's parameters and input data become aligned leading to Law of Large Numbers (LLN) scaling, then the update, initialization, and pre-activation multiplier above is sufficient to obtain a Maximal Update Parametrization. }
\begin{proof}
We will now prove both statements by arguing that, in each case, the update of $f_\phi$ is in $\Theta(1)$, implying that our parameterization is correct. Without loss of generality, we will assume that the optimizee network has input dimension $d$, hidden dimension $n$ (width), and output dimension $c$. Let $\mW$ be some weight matrix in the optimizee network, let the update produced by $f_\phi$ be $\Delta \mW$ and let $\mA$ be the corresponding input features such that $\Delta \mW=f_\phi(\mA)$. 
\begin{itemize}
    \item In the case of $\texttt{small\_fc\_lopt}$, $f_\phi(\vx) = \Theta(1)$ since its input features, $\mA$, are $\Theta(1)$ due to normalization (see corollary~\ref{coro:smallfc}). 
    \item In the case of VeLO, we must also show that the LSTM hypernetwork does not introduce additional dependence on the width, $n$. From corollary~\ref{coro:velo} we know that the LSTM hypernetwork will produce parameters, $\phi_\mW$, of $\texttt{small\_fc\_lopt}$ and an LR multiplier, $\alpha_\mW$ which are $\Theta(1)$ since all inputs to the LSTM are $\Theta(1)$. Therefore, $f_\phi(\vx) = \Theta(1)$ for VeLO aswell. 
\end{itemize}
This fact is henceforth referred to as property (A). We will assume that the optimizee network follows our proposed $\mu$-parameterization from Sec.~\ref{sec:proof}, and show that we satisfy the desiderata of $\mu$P (outlined above) for any weight layer, $\mW$, in the network. Concretely, we will show that for input and hidden layers,
\begin{align}\label{eq:mup-hidden}
   \vx_i = \Theta(1) \Rightarrow (\mW \vx)_i = \Theta(1) \text{ and } ((\mW + \Delta \mW)\vx)_i = \Theta(1)
\end{align}
that for the output layer
\begin{align}\label{eq:mup-input}
    \vx_i = \Theta(1) \Rightarrow (\mW \vx)_i = O(1) \text{ and } ((\mW + \Delta \mW)\vx)_i = O(1)
\end{align}
and that for all layers
\begin{align}\label{eq:maximalupdates}
   (\Delta \mW\vx)_i = \Theta(1).
\end{align}
Statements~\ref{eq:mup-hidden}, ~\ref{eq:mup-input}, and~\ref{eq:maximalupdates} correspond to desiderata (1) activation scale, (2) output scale, and (3) maximal updates, respectively. 
 In satisfying these statements, we will show that our parameterization is indeed a maximal update parameterization.

\paragraph{Output weights.}
Here, the input $\vx$ has $\Theta(1)$ coordinates, we initialize the output matrix $\mW$ with entries of variance 1 (which is necessary) and rescale the logits with $1/n$. Therefore, the output, $(1/n)\mW\vx$, is $O(1)$ by the LLN \textbf{(Output Scale Property)}. From property (A), we know that $\Delta \mW=f_\phi(\nabla \mW)$ has coordinates in $\Theta(1)$, so the entries of $\mW + \Delta \mW$ still have variance $1$ and $\tfrac1n((\mW + \Delta \mW)\vx)_i$ is $O(1)$. Moreover, $\frac{1}{n}(\Delta \mW x)_i = \Theta(1)$ by LLN (\textbf{Maximal updates}).

\paragraph{Hidden weights.} Since hidden weights are initialized with variance $1/n$ and $\vx_i = \Theta(1)$, the coordinates of $\mW\vx$ are $\Theta(1)$ by LLN. From property (A), we know that $f_\phi(\mA) = \Theta(1)$. Therefore, to ensure $\Delta \mW \cdot \vx$ is coordinate-wise bounded, we must re-scale the parameter updates:
\[
\Delta \mW = \frac{1}{n} f_\phi(\mA).
\]
Since this rescaling implies that $\Delta \mW$ is $\Theta(1/n)$, the entries of $\mW + \Delta \mW$ still scale like $1/n$ and $((\mW + \Delta \mW)\vx)_i$ is $\Theta(1)$. Moreover, since $\Delta \mW$ is $\Theta(1/n)$ and $\vx_i = \Theta(1)$, then $1/n(\Delta \mW x)_i = \Theta(1)$ by LLN (\textbf{Maximal updates}).

\paragraph{Input weights.}
Recall that $d$, the input dimension, is fixed and does not grow with $n$. Since the input $\vx_i = \Theta(1)$ and $\mW$ has entries with variance $1/d$ in $\Theta(1)$, then the coordinates of pre-activation $\mW x$ are $\Theta(1)$. From property (A), we know that $f_\phi(\mA) = \Theta(1)$. Therefore, $\Delta \mW$ is $\Theta(1)$, the entries of $\mW + \Delta \mW$ still have $\Theta(1)$ coordinates  and $((\mW + \Delta \mW)\vx)_i$ is $\Theta(1)$ (as $d$ is fixed). Moreover, $\Delta \mW x$ will have coordinate sizes that depend on the input dimension, $d$, but not the width. Therefore, $(\Delta W x)_i = \Theta(1)$(\textbf{Maximal updates}).
\end{proof}

\subsection{Summary of learned optimizer input features}
The following section contains easy-to-read tables which report the exact learned optimizer input features for \texttt{small\_fc\_lopt} (Table~\ref{apdx:table:features-lopt}) and \texttt{VeLO} (Tables~\ref{apdx:table:features-velo-adafac} and~\ref{apdx:table:features-velo-lstm}). The tables also report the entry-wise scaling of the features before RMS-normalization and the number of features of each type. Entry-wise scaling is reported assuming a hidden weight matrix. The original implementation of these optimizers along with features calculation can be accessed here\footnote{\url{https://github.com/google/learned_optimization/blob/main/learned_optimization/learned_optimizers/adafac_mlp_lopt.py} and \url{https://github.com/google/learned_optimization/blob/main/learned_optimization/research/general_lopt/hyper_v2.py}}.

\renewcommand\tabularxcolumn[1]{m{#1}} 
\begin{table*}[ht]
\small
\begin{center}
    \caption{\textbf{$\mu$P scaling for hidden layers of per-parameter features input to $\mu$LO$_M$.} All the coefficients, $\beta_i$, are learnable parameters adjusted during meta-optimization. All feature calculations and scalings are reported for a hidden weight matrix $\mW \in \R^{m\times n}$ in an optimizee network following our proposed $\mu$-parameterization. Here, $n$ is the width and $m=kn$ for some constant $k\in\R$. In this case, the entries of the gradient of $\mW$, $\nabla_t$, scale like $\Theta(\frac{1}{n})$, where $n$ is the width of the model. \textbf{Notation.} The table will use $\nabla_{t,i}$ or $\nabla_{t,j}$ to indicate the variable's dependence on time $t$ and coefficient $\beta_i$ or $\beta_j$, respectively. $(\nabla_{t,j})_{r,c}$ will designate indexing into row $r$ and column $c$ of the quantity $\nabla_{t,j}$. \textbf{DISCLAIMER: All features in our tables report scaling before the RMS-normalization.}   }
    \label{apdx:table:features-lopt}
    \begin{tabularx}{\textwidth}{llX|l|cc}
        \toprule
        \multicolumn{1}{c}{\textbf{Type}}&\multicolumn{1}{c}{\textbf{\#}}&\multicolumn{1}{c|}{\textbf{Description}} &  \multicolumn{1}{c|}{\textbf{Accumulator Update/Equation}}  & \textbf{Scaling}\\  
        \midrule
        \multirow{9}{*}{\textbf{Accumulators}} 
        &3&Momentum accumulators with coefficients $\beta_i, i\in\{1,2,3\}$.                         & $\mM_{t, i} = \beta_i \mM_{t-1, i} + (1 - \beta_i) \nabla_t $                                                    & $\Theta(\frac{1}{n})$ \\[3mm]
        &1& Second moment accumulator with coefficient $\beta_4$.                                   & $\mV_t = \beta_4 \mV_{t-1} + (1 - \beta_4) \nabla_t^2 $                                                          & $\Theta(\frac{1}{n^2})$ \\[3mm]
        &3& Adafactor row accumulator with coefficients $\beta_i, i\in\{5,6,7\}$.                    & $\vr_{t, i} = \beta_i \vr_{t-1, i} + (1 - \beta_i)~\texttt{row\_mean}(\nabla_t^2) $                                 & $\Theta(\frac{1}{n^2})$\\[3mm]
        &3& Adafactor accumulator with coefficients $\beta_i, i\in\{5,6,7\}$.                        & $\vc_{t, i} = \beta_i \vc_{t-1, i} + (1 - \beta_i)~\texttt{col\_mean}(\nabla_t^2) $                                & $\Theta(\frac{1}{n^2})$\\\midrule
        \multirow{12}{*}{\parbox{2cm}{\textbf{Accumulator} \\ \textbf{~~~Features}}} 
        &3& Momentum values normalized by the square root of the second moment for $i\in\{5,6,7\}$.  & $\displaystyle\frac{\mM_{t,i}}{\sqrt{\mV_t}}$                                                                    & $\Theta(1)$ \\[3mm]
        &1&The reciprocal square root of the second moment value.                                    & $\displaystyle\frac{1}{\sqrt{\mV}}$                                                                              & $\Theta(n)$ \\[3mm]
        &6&The reciprocal square root of the Adafactor accumulators.                                 & $\displaystyle\frac{1}{\sqrt{\vr_{t,i}}} \textsc{  or  } \frac{1}{\sqrt{\vc_{t,i}}}$                             & $\Theta(n)$\\[3mm]
        &3& Adafactor gradient features for $i\in\{5,6,7\}$.                                         & $\displaystyle\nabla_t \cdot \sqrt{\frac{\frac{1}{m}\sum_{h=1}^m(\vr_{t, i})_h}{ \vr_{t, i} \vc_{t, i}^T}}$      & $\Theta(1)$ \\[3mm]
        &3& Adafactor momentum features for $i,j\in\{(5,1),(6,2),$ $(7,3)\}$.                        & $\displaystyle\mM_{t,j} \cdot \sqrt{\frac{\frac{1}{m}\sum_{h=1}^m(\vr_{t, i})_h}{ \vr_{t, i} \vc_{t, i}^T}} $    & $\Theta(1)$ \\\midrule
       \textbf{Time Features}&11& Time Features for $x \in \{ 1, 3, 10,$ $ 30, 100, 300, $ $1000, 3000, 10^4,$ $  3\cdot10^4, 10^5\}$.   &$\tanh{\left( \frac{t}{x} \right)}$                                           & $\Theta(1)$ \\\midrule
       \multirow{2}{*}{\textbf{Parameters}} &1&Parameter value.                                       & $\mW_{t}$                                                                                                       & $\Theta(\frac{1}{n})$\\
                                            &1&Gradient value.                                        & $\nabla_{t}$                                                                                                    &  $\Theta(\frac{1}{n})$\\\midrule
        \textbf{Total}&$39$&--&--&--\\
    \bottomrule
    \end{tabularx}
\end{center}
\end{table*}

\renewcommand\tabularxcolumn[1]{m{#1}} 
\begin{table*}[ht]
\small
\begin{center}
    \caption{\textbf{$\mu$P scaling of per-parameter features input to the per-parameter network of $\mu$VeLO$_M$.} All feature calculations and scalings are reported for a hidden weight matrix $\mW \in \R^{m\times n}$ in an optimizee network following our proposed $\mu$-parameterization. Here, $n$ is the width and $m=kn$ for some constant $k\in\R$. In this case, the entries of the gradient of $\mW$, $\nabla_t$, scale like $\Theta(\frac{1}{n})$, where $n$ is the width of the model. \textbf{Notation.} The table will use $\nabla_{t,i}$ or $\nabla_{t,j}$ to indicate the variable's dependence on time $t$ and coefficient $\beta_i$ or $\beta_j$, respectively. $(\nabla_{t,j})_{r,c}$ will designate indexing into row $r$ and column $c$ of the quantity $\nabla_{t,j}$. \textbf{DISCLAIMER: All features in our tables report scaling before the RMS-normalization.}  }
    \label{apdx:table:features-velo-adafac}
    \begin{tabularx}{\textwidth}{llX|l|cc}
        \toprule
        \multicolumn{1}{c}{\textbf{Type}}&\multicolumn{1}{c}{\textbf{\#}}&\multicolumn{1}{c|}{\textbf{Description}} &  \multicolumn{1}{c|}{\textbf{Accumulator Update/Equation}}  & \textbf{Scaling}\\  
        \midrule
        \multirow{13}{*}{\textbf{Accumulators}} 
        &3&Momentum accumulators with coefficients $\beta_i, i\in\{1,2,3\}$.                         & $\mM_{t, i} = \beta_i \mM_{t-1, i} + (1 - \beta_i) \nabla_t $                                                    & $\Theta(\frac{1}{n})$ \\[3mm]
        &1& Second moment accumulator with coefficient $\beta_4$.                                   & $\mV_t = \beta_4 \mV_{t-1} + (1 - \beta_4) \nabla_t^2 $                                                          & $\Theta(\frac{1}{n^2})$ \\[3mm]
        &3& Adafactor row accumulator with coefficients $\beta_i, i\in\{5,6,7\}$.                    & $\vr_{t, i} = \beta_i \vr_{t-1, i} + (1 - \beta_i)~\texttt{row\_mean}(\nabla_t^2) $                                 & $\Theta(\frac{1}{n^2})$\\[3mm]
        &3& Adafactor accumulator with coefficients $\beta_i, i\in\{5,6,7\}$.                        & $\vc_{t, i} = \beta_i \vc_{t-1, i} + (1 - \beta_i)~\texttt{col\_mean}(\nabla_t^2) $                                & $\Theta(\frac{1}{n^2})$\\\midrule
        \multirow{17}{*}{\parbox{2cm}{\textbf{Accumulator} \\ \textbf{~~~Features}}} 
        &3& Momentum values normalized by the square root of the second moment for $i\in\{5,6,7\}$.  & $\displaystyle\frac{\mM_{t,i}}{\sqrt{\mV_t}}$                                                                    & $\Theta(1)$ \\[3mm]
        &1&The reciprocal square root of the second moment value.                                    & $\displaystyle\frac{1}{\sqrt{\mV}}$                                                                              & $\Theta(n)$ \\[3mm]
        &6&The reciprocal square root of the Adafactor accumulators.                                 & $\displaystyle\frac{1}{\sqrt{\vr_{t,i}}} \textsc{  or  } \frac{1}{\sqrt{\vc_{t,i}}}$                             & $\Theta(n)$\\[3mm]
        &3& Adafactor gradient features for $i\in\{5,6,7\}$.                                         & $\displaystyle\nabla_t \cdot \sqrt{\frac{\frac{1}{m}\sum_{h=1}^m(\vr_{t, i})_h}{ \vr_{t, i} \vc_{t, i}^T}}$      & $\Theta(1)$ \\[3mm]
        &3& Adafactor momentum features for $i,j\in\{(5,1),(6,2),$ $(7,3)\}$.                        & $\displaystyle\mM_{t,j} \cdot \sqrt{\frac{\frac{1}{m}\sum_{h=1}^m(\vr_{t, i})_h}{ \vr_{t, i} \vc_{t, i}^T}} $    & $\Theta(1)$ \\\midrule
       \multirow{2}{*}{\textbf{Parameters}} &1&Parameter value.                                       & $\mW_{t}$                                                                                                       & $\Theta(\frac{1}{n})$\\
                                            &1&Gradient value.                                        & $\nabla_{t}$                                                                                                    &  $\Theta(\frac{1}{n})$\\
                                            &1&Gradient value.                                        & $\texttt{clip}(\nabla_{t},-0.1,0.1)$                                                                                                    &  $\Theta(\frac{1}{n})$\\\midrule
        \textbf{Total}&$29$&--&--&--\\
    \bottomrule
    \end{tabularx}
\end{center}
\end{table*}

\begin{table*}[ht]
\small
\begin{center}
    \caption{\textbf{Per-tensor features used as input to VeLO's LSTM.}  All feature calculations and scalings are reported for a hidden weight matrix $\mW \in \R^{m\times n}$ in an optimizee network following our proposed $\mu$-parameterization. Here, $n$ is the width and $m=kn$ for some constant $k\in\R$. In this case, the entries of the gradient of $\mW$, $\nabla_t$, scale like $\Theta(\frac{1}{n})$, where $n$ is the width of the model. }
    \label{apdx:table:features-velo-lstm}
    \begin{tabularx}{\textwidth}{llX|ll}
        \toprule
        \multicolumn{1}{c}{\textbf{Type}}&\multicolumn{1}{c}{\textbf{\#}}&\multicolumn{1}{c|}{\textbf{Description}} &  \multicolumn{1}{c}{\textbf{Equation}}&  \multicolumn{1}{c}{\textbf{Scaling}}\\  
        \midrule
        \multirow{2}{*}{\parbox{1.5cm}{\textbf{Accumulator} \\ \textbf{~~~Features}}} 
        &3& Variance across coordinates of the 3 momentum accumulator matrices normalized by the RMS of the current parameter values $i\in\{1,2,3\}$    & $\texttt{var\_mom}_i$ (Sec.~\ref{apdx:sec:velo})  & $\Theta(1)$ \\
        &1& Mean across coordinates of variance accumulator normalized by the parameter RMS                                                             & $\texttt{mean\_rms}$ (Sec.~\ref{apdx:sec:velo})  & $\Theta(1)$ \\
        &3& Coordinate-wise mean of the variance accumulator. $i\in\{1,2,3\}$                                                                           & $\texttt{var\_rms}_i$(Sec.~\ref{apdx:sec:velo})& $\Theta(1)$ \\
       \textbf{Tensor Rank}
       &5& A one hot vector representing the tensor's rank, $r$.                                                                                             &$e_r$              & $\Theta(1)$ \\\midrule
       \textbf{EMA Loss Features}
       &9&EMAs of the loss at different timescales chosen based on the number of steps. Values are normalized by the max and min losses seen so far. & see~\cite{velo}               & $\Theta(1)$ \\\midrule
       \multirow{1}{*}{\parbox{2cm}{\vspace{-3pt}\textbf{Remaining} \\ \textbf{Time Features}}}
       &9& Time Features for $x \in \{ 0.03, 0.1, 0.2, 0.4, $ $0.6, 0.8, 0.9, 1.0, 1.1\}$.                                                              &$\tanh( t/T - 10x)$     & $\Theta(1)$ \\\midrule
        \textbf{Total}&$30$&--&--&--\\
    \bottomrule
    \end{tabularx}
\end{center}
\end{table*}
\clearpage



\begin{table}[t]
\centering
\caption{\textbf{Meta-training and hyperparameter configurations of LOs and baselines in our empirical evaluation.} The small\_fc\_lopt and VeLO architectures were initially proposed in~\citep{metz2022practical} and ~\citep{velo}. See Tab.~\ref{apdx:tab:testtasks} for a list of all tasks used in this work.}
\label{tab:optimizers}
\begin{adjustbox}{width=\columnwidth,center}
\begin{tabular}{l|llll}
\toprule

\textbf{Identifier} & \textbf{Type} & \textbf{Architecture}&\textbf{Optimizee Par.} & \textbf{Meta-Training / Tuning Task(s)} \\
\midrule
  $\mulos$ & Ours & small\_fc\_lopt& $\mu$LO Sec.~\ref{sec:proof}   & IN32$\gT_{(3,128)}^\text{MLP}$ \\
  $\mulom$ & Ours & small\_fc\_lopt& $\mu$LO Sec.~\ref{sec:proof}   & IN32$\gT_{(3,128)}^\text{MLP}$,IN32$\gT_{(3,512)}^\text{MLP}$,IN32$\gT_{(3,1024)}^\text{MLP}$ \\
  $\mu$VeLO$_M$& Ours & VeLO& $\mu$LO Sec.~\ref{sec:proof}   & IN32$\gT_{(3,128)}^\text{MLP}$,IN32$\gT_{(3,512)}^\text{MLP}$,IN32$\gT_{(3,1024)}^\text{MLP}$\\\midrule
  $\los$ & LO Baseline &small\_fc\_lopt& SP  & IN32$\gT_{(3,128)}^\text{MLP}$ \\
  $\lom$ & LO Baseline & small\_fc\_lopt& SP  & IN32$\gT_{(3,128)}^\text{MLP}$,IN32$\gT_{(3,512)}^\text{MLP}$,IN32$\gT_{(3,1024)}^\text{MLP}$ \\
  VeLO$_M$ & LO Baseline & VeLO&  SP  & IN32$\gT_{(3,128)}^\text{MLP}$,IN32$\gT_{(3,512)}^\text{MLP}$,IN32$\gT_{(3,1024)}^\text{MLP}$ \\
  VeLO-4000 & Oracle LO Baseline &VeLO& SP  & See \cite{velo} (Appendix C.2)\\\midrule
  $\mu$Adam & Baseline & -- & $\mu$P Adam   & per-task tuning (see Tab.~\ref{apdx:tab:muadamoptimal})\\
  AdamW & Baseline & -- & SP  & per-task tuning (see Tab.~\ref{apdx:tab:adamwoptimal})\\ 
\bottomrule
\end{tabular}
\end{adjustbox}
\end{table}

\section{Hand Designed Optimizer Hyperparameter Tuning}\label{apdx:sec:hp-tuning}
To provide strong baselines for our study, we tune the hyperparameters of AdamW and $\mu$Adam for more than $500$ trials on one instance of each task in our evaluation suite. Since the largest width task seen by $\mulom$ and $\muvelom$ is $1024$, we select this width for all our hyperparameter sweeps. Similarly, we use the same depth=$3$ and training steps=$1000$ as for the meta-training of $\mulom$ and $\muvelom$.

\subsection{Tuning $\mu$Adam}
\label{sec:apdx:muadam-tuning}
We tune $\mu$Adam's learning rate ($\eta$) and accumulator coefficients ($\beta_1$, and $\beta_2$). Table~\ref{apdx:tab:muadamhps} reports all hyperparameter values that we swept for each task. Table~\ref{apdx:tab:muadamoptimal} reports the best-performing hyperparameter values found by selecting the values that achieved the lowest final smoothed training loss on each task. 

\begin{table}[h]
    \centering
    \caption{\textbf{Hyperparameter sweep values for $\mu$Adam.}}
    \label{apdx:tab:muadamhps}
    \begin{adjustbox}{width=1.0\columnwidth,center}
    \begin{tabular}{cc|l}
        \toprule
        Hyperparameter & \textbf{\#} & Values \\
        \midrule
        $\eta$ &32& $\{10^{-6}, 1.56 \times 10^{-6}, 2.44 \times 10^{-6}, 3.81 \times 10^{-6}, 5.95 \times 10^{-6},9.28 \times 10^{-6}, $
        \\ &&$1.45 \times 10^{-5}, 2.26 \times 10^{-5}, 3.53 \times 10^{-5}, 5.52 \times 10^{-5}, 8.62 \times 10^{-5},
        $ \\
        &&$1.35 \times 10^{-4}, 2.10 \times 10^{-4}, 3.28 \times 10^{-4}, 5.12 \times 10^{-4}, 8.00 \times 10^{-4},$ \\
        &&$ 1.25 \times 10^{-3}, 1.95 \times 10^{-3}, 3.05 \times 10^{-3}, 4.76 \times 10^{-3}, 7.43 \times 10^{-3},$ \\
        &&$ 1.16 \times 10^{-2}, 1.81 \times 10^{-2}, 2.83 \times 10^{-2}, 4.42 \times 10^{-2}, 6.90 \times 10^{-2},$ \\
        &&$ 1.08 \times 10^{-1}, 1.68 \times 10^{-1}, 2.63 \times 10^{-1}, 4.10 \times 10^{-1}, 6.40 \times 10^{-1}, 1\}$ \\
        $\beta_1$ &4& $\{0.85, 0.9, 0.95, 0.99\}$ \\
        $\beta_2$ &4& $\{0.9, 0.95, 0.99, 0.999\}$ \\\midrule
        \textbf{Total} &512&--\\
        \bottomrule
    \end{tabular}
\end{adjustbox}
\end{table}

\begin{table}[h]
    \centering
    \caption{\textbf{Strongest performing hyperparameter values of $\mu$Adam for each task, with and without a schedule.} All optimizers with a schedule use a linear warmup and cosine decay schedule with the minimum learning rate set to $0.1\eta$. }
    \label{apdx:tab:muadamoptimal}
    \begin{tabular}{l | c c c | c }
        \toprule
        \textbf{Task} & $\eta$ & $\beta_1$ & $\beta_2$ & GPU Hours\\
        \midrule
        \multirow{1}{*}{$\gT_{(3,1024)}^\text{LM}$}
            & 0.1077  & 0.85 & 0.999 & 48  \\
        \multirow{1}{*}{$\gT_{(3,1024)}^\text{ViT}$}
            & 0.044173 & 0.85 & 0.999 & 17  \\
        \multirow{1}{*}{IN32$\gT_{(3,1024)}^\text{MLP}$}
            & 0.044173 & 0.85 & 0.999 & 9  \\
        \multirow{1}{*}{IN64$\gT_{(3,1024)}^\text{MLP}$}
            & 0.028289 & 0.85 & 0.99  & 19  \\
        \multirow{1}{*}{C10$\gT_{(3,1024)}^\text{MLP}$}
            & 0.1473   & 0.9  & 0.95  & 4   \\
            \midrule
        \multirow{1}{*}{\textbf{Total}}
            & --  & --  & --  & 97   \\
        \bottomrule
    \end{tabular}
\end{table}

\subsection{Tuning AdamW}
\label{sec:apdx:adamw-tuning}
We tune AdamW's learning rate ($\eta$), accumulator coefficients ($\beta_1$, and $\beta_2$), and weight decay ($\lambda$). Table~\ref{apdx:tab:adamwhps} reports all hyperparameter values that we swept for each task. Table~\ref{apdx:tab:adamwoptimal} reports the best-performing hyperparameter values found by selecting the values that achieved the lowest final smoothed training loss on each task.

\begin{table}[h!]
    \centering
    \caption{\textbf{Hyperparameter sweep values for AdamW.}}
    \label{apdx:tab:adamwhps}
\begin{tabular}{cc|l}
    \toprule
    \textbf{Hyperparameter}&\textbf{\#} & \textbf{Values} \\
    \midrule
    $\eta$ &14 & $\{0.1,\ 4.92 \times 10^{-2},\ 2.42 \times 10^{-2},\ 1.19 \times 10^{-2},$ \\
          && \quad $5.88 \times 10^{-3}, 2.89 \times 10^{-3},\ 1.43 \times 10^{-3}, $ \\
          && \quad $7.02 \times 10^{-4},\ 3.46 \times 10^{-4},\ 1.70 \times 10^{-4},$\\
          && \quad $8.38 \times 10^{-5},\ 4.12 \times 10^{-5},\ 2.03 \times 10^{-5},\ 1.00 \times 10^{-5}\}$ \\
    $\beta_1$ &3& $\{0.9,\ 0.95,\ 0.99\}$ \\
    $\beta_2$ &3& $\{0.95,\ 0.99,\ 0.999\}$ \\
    $\lambda$ &4& $\{0.1,\ 0.01,\ 0.001,\ 0.0001\}$ \\\midrule
    \textbf{Total}&504&--\\
    \bottomrule
\end{tabular}
\end{table}

\begin{table}[h!]
    \centering
    \caption{\textbf{Strongest performing hyperparameter values of AdamW for each task, with and without a schedule.} All optimizers with a schedule use a linear warmup and cosine decay schedule with the minimum learning rate set to $0.1\eta$. }
    \label{apdx:tab:adamwoptimal}
    \begin{tabular}{l | c c c c | c}
        \toprule
        \textbf{Task} & $\eta$ & $\beta_1$ & $\beta_2$ & $\lambda$ & GPU Hours\\
        \midrule
        \multirow{1}{*}{$\gT_{(3,1024)}^\text{LM}$} 
            & $7.02 \times 10^{-4}$ & 0.9 & 0.99  & 0.001 & 48  \\
        \multirow{1}{*}{$\gT_{(3,1024)}^\text{ViT}$} 
            & $1.70 \times 10^{-4}$ & 0.9 & 0.999 & 0.01  & 18  \\
        \multirow{1}{*}{IN32$\gT_{(3,1024)}^\text{MLP}$} 
            & $7.02 \times 10^{-4}$ & 0.9 & 0.999 & 0.01  & 9   \\
        \multirow{1}{*}{IN64$\gT_{(3,1024)}^\text{MLP}$} 
            & $7.02 \times 10^{-4}$ & 0.9 & 0.99  & 0.001 & 20  \\
        \multirow{1}{*}{C10$\gT_{(3,1024)}^\text{MLP}$}  
            & $2.89 \times 10^{-3}$ & 0.9 & 0.95  & 0.0001& 4  \\
         \midrule
        \multirow{1}{*}{\textbf{Total}}
            & --  & --  & --  & -- & 99   \\
        \bottomrule
    \end{tabular}
\end{table}

\clearpage
\section{Meta-training with $\mu$LOs}\label{apdx:sec:meta-train}
\begin{figure}[h]
    \centering
    \includegraphics[width=\linewidth]{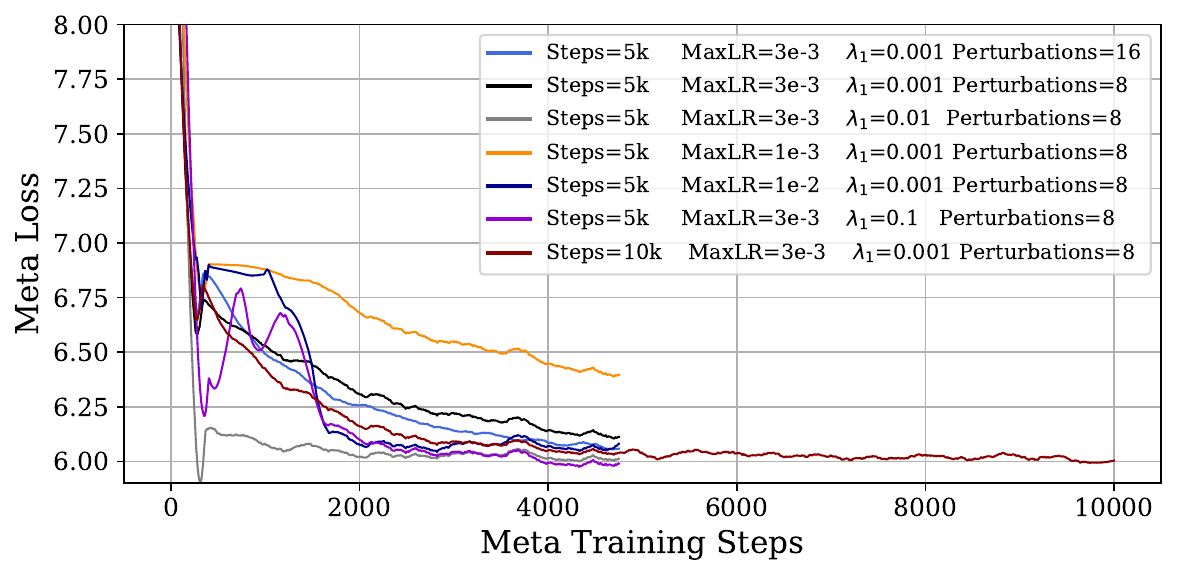}
    \caption{\textbf{Ablating Meta-training Hyperparameter for $\mulos$.} All curves show a single meta-training run. Using AdamW with a linear warmup and cosine annealing schedule, we meta-train $\mulos$ to train 3-layer width 128 MLPs for classifying $32\times 32\times3$ ImageNet Images. By default, we warmup linearly for 100 steps to a maximum learning rate of $3e-3$ and anneal the learning rate for $4900$ steps to a value of $1e-3$ with $\lambda_1=0.001$ (from Equation \ref{eq:update}) and sampling $8$ perturbations per step in PES\cite{vicol2021pes}. The above ablation varies the maximum learning rate $\in\{1e-2,3e-3,1e-3\}$ (always using $100$ steps of warmup and decaying to $0.3\times$MaxLR), $\lambda_1 \in \{0.001,0.01,0.1\}$, the number of steps (5k or 10k), and the number of perturbations (8 or 16). We observe that using all default values except for $\lambda_1=0.01$ yields one of the best solutions while being fast to train and stable during meta-training.}
    \label{fig:meta-train-ablation}
\end{figure}
\paragraph{General meta-training setup for small\_fc\_lopt} Each small\_fc\_lopt~\citep{metz2022practical} learned optimizer is meta-trained for $5000$ steps of gradient descent using AdamW \citep{adamw} and a linear warmup and cosine annealing schedule. We use PES~\citep{vicol2021pes} to estimate meta-gradients with a truncation length of $50$ steps and sampling $8$ perturbations per task at each step with standard deviation $0.01$. For the inner optimization task, we used a maximum unroll length of $1000$ iterations; that is, all our learned optimizers see at most $1000$ steps of the inner optimization problem during meta-training. Unlike with $\mu$Adam, we do not tune the \mup multipliers when meta-training $\mulos$ and $\mulom$, instead, we set them all to $1$. All optimizers are meta-trained on a single A6000 GPU. $\mulos$ and $\los$ take $8$ hours each to meta-train, while $\mulom$ and $\lom$ take $103$ hours.

\paragraph{General meta-training setup for VeLO} Each VeLO~\citep{metz2022practical} learned optimizer is meta-trained for $45000$ steps of gradient descent using AdamW~\citep{adamw} and a linear warmup and cosine annealing schedule. We using PES~\citep{vicol2021pes} to estimate meta-gradients with a truncation length of $20$ steps and sampling $8$ perturbations per task at each step with standard deviation $0.01$. For the inner optimization task, we used a maximum unroll length of $1000$ iterations; that is, all our learned optimizers see at most $1000$ steps of the inner optimization problem during meta-training. Unlike~\cite{tensorv}, we do not tune the \mup multipliers when meta-training $\mulos$ and $\mulom$, instead, we set them all to $1$. All optimizers are meta-trained on a single A6000 GPU. $\mu$VeLO$_M$and VeLO$_M$ each take $250$ GPU-hours to meta-train.

\paragraph{Meta-training hyperparameters for small\_fc\_lopt in $\mu$P} While there are very few differences between $\mu$LOs and SP LOs, the effective step size for hidden layers is changed (see eq.~\ref{eq:update}) which could alter the optimal meta-training hyperparameters. Consequently, we conduct an ablation study on hyper-parameters choices for $\mulos$. Specifically, using AdamW and gradient clipping with a linear warmup and cosine annealing LR schedule, we meta-train $\mulos$ to train 3-layer width 128 MLPs to classify $32\times 32\times3$ ImageNet Images. By default, we warmup linearly for 100 steps to a maximum learning rate of $3e-3$ and anneal the learning rate for $4900$ steps to a value of $1e-3$ with $\lambda_1=0.001$ (from Equation \ref{eq:update}) and sampling $8$ perturbations per step in PES\cite{vicol2021pes}. The above ablation varies the maximum learning rate $\in\{1e-2,3e-3,1e-3\}$ (always using $100$ steps of warmup and decaying to $0.3\times$MaxLR), $\lambda_1 \in \{0.001,0.01,0.1\}$, the number of steps (5k or 10k), and the number of perturbations (8 or 16). We observe that using all default values except for $\lambda_1=0.01$ yields one of the best solutions while being fast to train and stable during meta-training. We, therefore, select these hyperparameters to meta-train $\mulos$ and $\mulom$.

\paragraph{Meta-training hyperparameters for VeLO in $\mu$P} Unlike for small\_fc\_lopt, we do not find it necessary to change $\lambda_1$ from its default value of $0.001$. However, we do slightly alter the VeLO update by removing the multiplication by the current parameter norm. This causes problems when initializing tensors to zero, as we do in our experiments.

\paragraph{\mup at Meta-training time} It is important to carefully choose meta-training tasks that can effectively be transferred to larger tasks. In \citep{tensorv}, authors discuss these points and provide two notable guidelines: initialize the output weight matrix to zero (as it will approach zero in the limit) and use a relatively large key size when meta-training transformers. For all our tasks, we initialize the network's final layer to zeros following this guidance. While we do not meta-train on transformers, we suspect that the aforementioned transformer-specific guidelines may be useful for doing so.

\section{Extended Related Work}
\paragraph{Learned optimization.}
While research on learned optimizers (LOs) spans several decades~\citep{schmidhuber1992learning,thrun2012learning,chen2022learning,amos2022tutorial}, our work is primarily related to the recent meta-learning approaches utilizing efficient per-parameter optimizer architectures of~\cite{metz2022practical}. Unlike prior work~\citep{andrychowicz2016learning,wichrowska2017learned,chen2020training}, which computes meta-gradients (the gradients of the learned optimizer) using backpropagation, \cite{metz2022practical} use Persistent Evolutionary Strategies (PES)~\citep{vicol2021pes}, a truncated variant of evolutionary strategies (ES)~\citep{buckman2018es,nesteroves,parmas2018es}. ES improves meta-training of LOs by having more stable meta-gradient estimates compared to backpropagation through time, especially for longer sequences (i.e. long parameter update unrolls inherent in meta-training)~\citep{metz2019understanding}. PES and most recently ES-Single~\citep{vicol2023low} are more efficient and accurate variants of ES, among which PES is more well-established in practice making it a favourable approach to meta-training.
\paragraph{Generalization in LOs.}
One of the critical issues in LOs is generalization in the three main aspects~\citep{chen2022learning,amos2022tutorial}: (1) optimize novel tasks (often referred to as \textit{meta-generalization}); (2) optimize for more iterations than the maximum unroll length used in meta-training; (3) avoid overfitting on the training set.
Among these, (3) has been extensively addressed using different approaches, such as meta-training on the validation set objective~\citep{metz2019understanding}, adding extra-regularization terms~\citep{harrison2022closer}, parameterizing LOs as hyperparameter controllers~\citep{almeida2021generalizable} and introducing flatness-aware regularizations~\citep{yang2023learning}.
The regularization terms~\citep{harrison2022closer,yang2023learning} often alleviate issue (2) as a byproduct. 
However, meta-generalization (1) has remained a more difficult problem.

One approach to tackle this problem is to meta-train LOs on thousands of tasks~\citep{velo}. However, this approach is extremely expensive and does not address the issue in a principled way leading to poor meta-generalization in some cases, e.g. when the optimization task includes much larger networks.
Alternatively, \cite{premont2022simple} introduced Loss-Guarded L2O (LGL2O) that switches to Adam/SGD if the LO starts to diverge improving meta-generalization. However, this approach needs tuning Adam/SGD and requires additional computation (e.g. for loss check) limiting (or completely diminishing in some cases) the benefits of the LO. In this work, we study aspects (1) and (2) of LO generalization, demonstrating how existing SP LOs generalize poorly across these dimensions and showing how one can apply \mup to learned optimizers to substantially improve generalization in both these aspects. 
\paragraph{Maximal Update Parametrization.}
First proposed by \cite{tensorprogramsiva}, the Maximal Update Parametrization is the unique stable abc-Parametrization where every layer learns features. The parametrization was derived for adaptive optimizers by \cite{tesnorprogramsivb} and was applied by \cite{tensorv} to enable zero-shot hyperparameter transfer, constituting the first practical application of the tensor programs series of papers. Earlier works in the \emph{tensor programs series} build the mathematical foundation that led to the discovery of \mup. \cite{tensorprogramsi} shows that many modern neural networks with randomly initialized weights and biases are Gaussian Processes, providing a language, called Netsor, to formalize neural network computations. \cite{tensorprogramsii} focuses on neural tangent kernels (NTK), proving that as a randomly initialized network's width tends to infinity, its NTK converges to a deterministic limit. \cite{tensorprogramsiii} shows that randomly initialized network's pre-activations become independent of its weights when its width tends to infinity. Most recently, in tensor programs VI, \cite{mudepth} propose Depth-$\mu$P, a parameterization allowing for hyperparameter transfer in infinitely deep networks. However, Depth-$\mu$P is only valid for residual networks with a block depth of $1$, making it unusable for most practical architectures (e.g., transformers, resnets, etc.). For these reasons, we do not study Depth-$\mu$P herein. Building on the latest works studying width $\mu$P~\citep{tesnorprogramsivb,tensorv}, in this work, we show that \mup can be extended to the case of learned optimizers and empirically evaluate its benefits in this setting.

\section{List of Meta-testing Tasks}\label{apdx:sec:tasks}
Table~\ref{apdx:tab:testtasks} reports the configuration of different testing tasks used to evaluate our optimizers. We note that we do not augment the ImageNet datasets we use in any way except for normalizing the images. We tokenize LM1B using a sentence piece tokenizer\citep{kudo2018sentencepiece} with 32k vocabulary size. All evaluation tasks are run on A6000 $48$GB or A100 $80$GB GPUs for $5$ random seeds.

\begin{table}[h]
\centering
\caption{\textbf{Meta-testing settings.} We report the optimization tasks we will use to evaluate the LOs of Table~\ref{tab:optimizers}. }
\label{apdx:tab:testtasks}
\begin{adjustbox}{width=\columnwidth,center}
\begin{tabular}{c|llllllll}
\toprule
\textbf{Identifier} & \textbf{Dataset} & \textbf{Model} &\textbf{Depth}& \textbf{Width} & \textbf{Attn. Heads} & \textbf{FFN Size}& \textbf{Batch Size} & \textbf{Sequence Length} \\
\midrule
IN32$\gT_{(3,128)}^\text{MLP}$ & $32\times32\times3$ ImageNet &  MLP & $3$ & $128$& -- & -- & $4096$ & -- \\
IN32$\gT_{(3,256)}^\text{MLP}$ & $32\times32\times3$ ImageNet &  MLP & $3$ & $256$& -- & -- & $4096$ & -- \\
IN32$\gT_{(3,512)}^\text{MLP}$ & $32\times32\times3$ ImageNet &  MLP & $3$ & $512$& -- & -- & $4096$ & -- \\
IN32$\gT_{(3,1024)}^\text{MLP}$ & $32\times32\times3$ ImageNet &  MLP & $3$ & $1024$& -- & -- & $4096$ & -- \\
IN32$\gT_{(3,2048)}^\text{MLP}$ & $32\times32\times3$ ImageNet &  MLP & $3$ & $2048$& -- & -- & $4096$ & -- \\
IN32$\gT_{(3,4096)}^\text{MLP}$ & $32\times32\times3$ ImageNet &  MLP & $3$ & $4096$& -- & -- & $4096$ & -- \\
IN32$\gT_{(3,8192)}^\text{MLP}$ & $32\times32\times3$ ImageNet &  MLP & $3$ & $8192$& -- & -- & $4096$ & -- \\
\midrule
IN64$\gT_{(3,128)}^\text{MLP}$ & $64\times64\times3$ ImageNet &  MLP & $3$ & $128$& -- & -- & $4096$ & -- \\
IN64$\gT_{(3,256)}^\text{MLP}$ & $64\times64\times3$ ImageNet &  MLP & $3$ & $256$& -- & -- & $4096$ & -- \\
IN64$\gT_{(3,512)}^\text{MLP}$ & $64\times64\times3$ ImageNet &  MLP & $3$ & $512$& -- & -- & $4096$ & -- \\
IN64$\gT_{(3,1024)}^\text{MLP}$ & $64\times64\times3$ ImageNet &  MLP & $3$ & $1024$& -- & -- & $4096$ & -- \\
IN64$\gT_{(3,2048)}^\text{MLP}$ & $64\times64\times3$ ImageNet &  MLP & $3$ & $2048$& -- & -- & $4096$ & -- \\
IN64$\gT_{(3,4096)}^\text{MLP}$ & $64\times64\times3$ ImageNet &  MLP & $3$ & $4096$& -- & -- & $4096$ & -- \\
\midrule
C10$\gT_{(3,128)}^\text{MLP}$ & $32\times32\times3$ Cifar-10 &  MLP & $3$ & $128$& -- & -- & $4096$ & -- \\
C10$\gT_{(3,256)}^\text{MLP}$ & $32\times32\times3$ Cifar-10 &  MLP & $3$ & $256$& -- & -- & $4096$ & -- \\
C10$\gT_{(3,512)}^\text{MLP}$ & $32\times32\times3$ Cifar-10 &  MLP & $3$ & $512$& -- & -- & $4096$ & -- \\
$\gT_{(3,1024)}^\text{LM}$ & $32\times32\times3$ Cifar-10 &  MLP & $3$ & $1024$& -- & -- & $4096$ & -- \\
C10$\gT_{(3,2048)}^\text{MLP}$ & $32\times32\times3$ Cifar-10 &  MLP & $3$ & $2048$& -- & -- & $4096$ & -- \\
C10$\gT_{(3,4096)}^\text{MLP}$ & $32\times32\times3$ Cifar-10 &  MLP & $3$ & $4096$& -- & -- & $4096$ & -- \\
C10$\gT_{(3,8192)}^\text{MLP}$ & $32\times32\times3$ Cifar-10 &  MLP & $3$ & $8192$& -- & -- & $4096$ & -- \\
\midrule
$\gT_{(3,192)}^\text{ViT}$ & $32\times32\times3$ ImageNet &  ViT & $3$ & $192$& $3$ & $768$ & $1024$ & -- \\
$\gT_{(3,384)}^\text{ViT}$ & $32\times32\times3$ ImageNet &  ViT & $3$ & $384$& $6$ & $1536$ & $1024$ & -- \\
$\gT_{(3,768)}^\text{ViT}$ & $32\times32\times3$ ImageNet &  ViT & $3$ & $768$& $8$ & $3072$ & $1024$ & -- \\
$\gT_{(3,1024)}^\text{ViT}$ & $32\times32\times3$ ImageNet &  ViT & $3$ & $1024$& $8$ & $4096$ & $1024$ & -- \\
$\gT_{(3,2048)}^\text{ViT}$ & $32\times32\times3$ ImageNet &  ViT & $3$ & $2048$& $16$ & $8192$ & $1024$ & -- \\
$\gT_{(3,3072)}^\text{ViT}$ & $32\times32\times3$ ImageNet &  ViT & $3$ & $3072$& $16$ & $12288$ & $1024$ & -- \\
\midrule
$\gT_{(3,192)}^\text{LM}$ & LM1B, $32k$ Vocab &   Transformer LM & $3$ & $192$& $3$ & $768$ & $128$ & 64 \\
$\gT_{(3,384)}^\text{LM}$ & LM1B, $32k$ Vocab &   Transformer LM & $3$ & $384$& $6$ & $1536$ & $128$ & 64 \\
$\gT_{(3,768)}^\text{LM}$ & LM1B, $32k$ Vocab &   Transformer LM & $3$ & $768$& $8$ & $3072$ & $128$ & 64 \\
$\gT_{(3,1024)}^\text{LM}$ & LM1B, $32k$ Vocab &   Transformer LM & $3$ & $1024$& $8$ & $4096$ & $128$ & 64 \\
$\gT_{(3,2048)}^\text{LM}$ & LM1B, $32k$ Vocab &   Transformer LM & $3$ & $2048$& $16$ & $8192$ & $128$ & 64 \\
$\gT_{(3,3072)}^\text{LM}$ & LM1B, $32k$ Vocab &   Transformer LM & $3$ & $3072$& $16$ & $12288$ & $128$ & 64 \\
\midrule
  $\gD\tmlp_{(16,1024)}$ & $32\times32$ ImageNet&  MLP & $16$ & $1024$& -- & -- & $128$ & -- \\
  $\gD\tvit_{(16,1024)}$ &  $32\times32$ ImageNet& ViT & $16$ & $1024$& $3$ & $4096$ & $128$ & -- \\
  $\gD\tlm_{(16,1024)}$ & LM1B& Transformer LM & $16$ & $1024$& $3$ & $4096$ & $128$ & -- \\
\bottomrule
\end{tabular}
\end{adjustbox}
\end{table}

\section{Additional Experiments}
\subsection{Comparison with VeLO-4000}

\noindent\textbf{Pre-trained VeLO (VeLO-4000).} VeLO~\citep{velo} is a learned optimizer that was meta-trained on a curriculum of progressively more expensive meta-training tasks for a total of $4000$ TPU months. These tasks include but are not limited to image classification with MLPs, ViTs, ConvNets, and ResNets; compression with MLP auto-encoders; generative modeling with VAEs; and language modeling with transformers and recurrent neural networks. During meta-training, VeLO-4000 unrolls inner problems for up to 20k steps ($20\times$ ours); the final model was then fine-tuned on tasks with up to 200k steps of optimization. VeLO-4000, therefore represents a strong but unfair baseline as it is trained on far more data and with far more compute than our main VeLO experiments. 

\noindent\textbf{Is VeLO-4000 a fair baseline?}
While we believe the comparison is interesting given the relevance of our results to scaling up LOs, the comparison will unfairly advantage VeLO-4000 as \textbf{all tasks in our suite fall within its meta-training distribution} and VeLO-4000 was meta-trained on inner unroll horizons well beyond those we evaluate. Thus, when comparing our LOs to VeLO-4000, it is important to keep in mind that it is an unfair baseline since our learned optimizers meta-trained with only $0.004\%$ of VeLO-4000's compute budget. We included a compute-matched fair baseline, $\velom$ in the main manuscript.

\begin{figure*}[h]
    \centering
     \subfloat[MLP IN32 W=$8192$ ]{\includegraphics[width=0.45\linewidth]{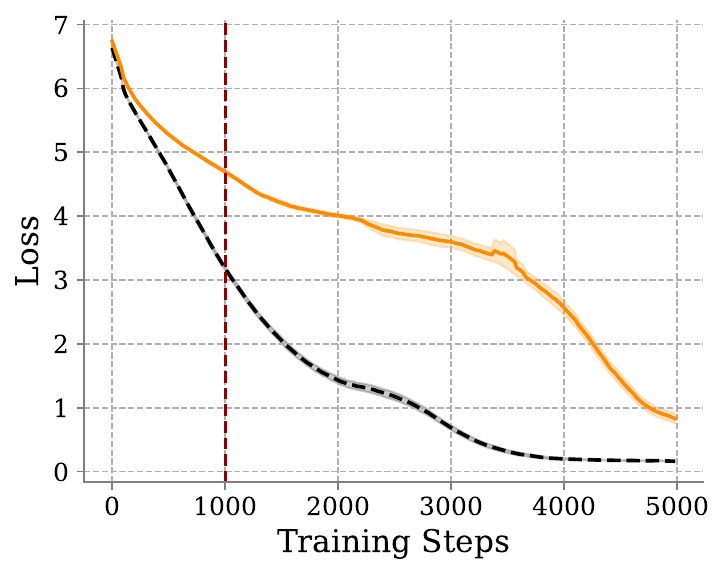}}\quad
    \subfloat[LM W=$4096$]{\includegraphics[width=0.45\linewidth]{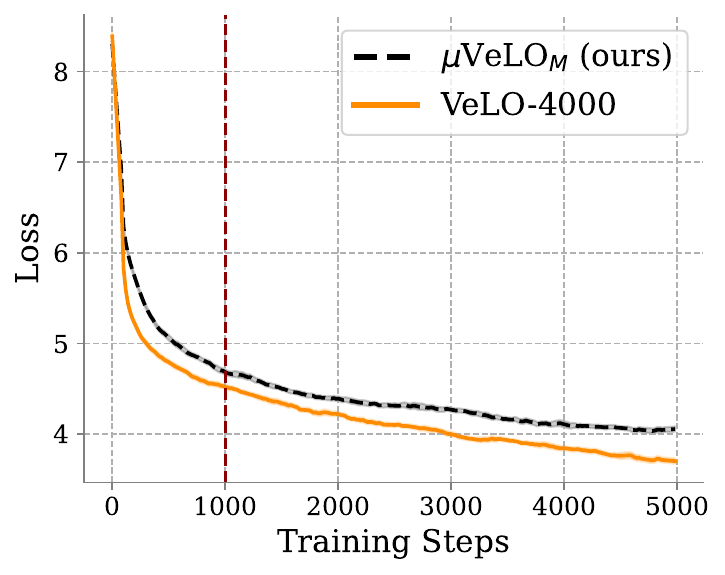}}
    \caption{\textbf{A comparison to VeLO-4000 on the widest tasks.} All optimizers except VeLO are meta-trained or hyperparameter tuned for $1000$ inner steps (dotted red line), therefore, any optimization beyond $1000$ steps is considered out-of-distribution. We plot average training loss over $5$ seeds with standard error bars. We observe that $\mulom$ and $\muvelom$ outperform VeLO on the widest in-distribution tasks, but fall short, despite still generalizing well when evaluated far out-of-distribution on a width $3072$ language modeling task.
    }
    \label{fig:results-velo-comp}
\end{figure*}

\noindent\textbf{Comparison} Figures~\ref{fig:results-velo-comp} reports the training curves of different optimizers, including VeLO-4000, on width $8192$ and $3072$ MLP and transformer language model tasks, respectively. We observe that $\mulom$ and $\muvelom$ (trained with many orders of magnitude less compute) outperforms VeLO-4000 at this large width on the in-distribution tasks, but fall short despite still generalizing well when evaluated far out-of-distribution on a width $3072$ language modeling task. We hypothesize that this is likely due to the task being nearly in-distribution for VeLO-4000 meta-training data while being OOD w.r.t. architecture, width, and training steps for $\mulom$ and $\muvelom$. These results overall suggest that $\muvelom$ may be more scalable than its non-$\mu$P counterpart, particularly in the large model cases where VeLO-4000 struggled \citep{velo}.

\clearpage
\subsubsection{Why do $\mu$LOs improve generalization to depth and longer training horizons?}\label{apdx:sec:explain}
\begin{figure}[h]
    \centering
    \subfloat[Increasing Depth]{\includegraphics[width=0.45\linewidth]{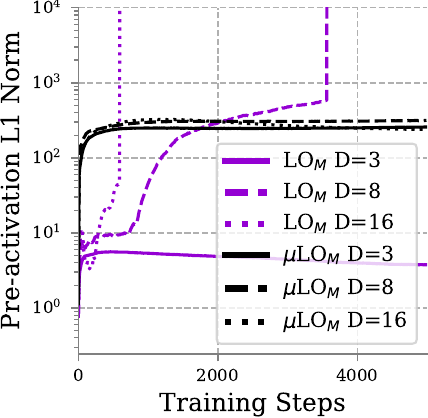}}\quad
    \subfloat[Longer Training]{\includegraphics[width=0.45\linewidth]{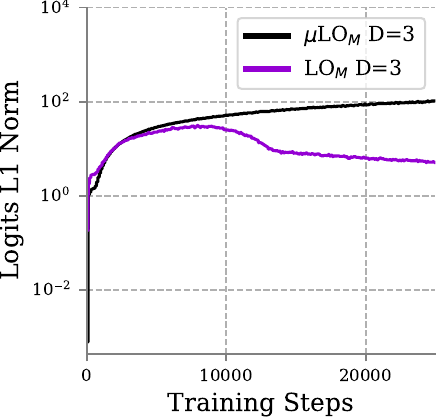}}
    \caption{\textbf{Activation stability for deeper and longer training.} Each curve reports the five-seed average L1 norm of first-layer pre-activation and logits for (a) and (b), respectively.}
    \label{apdx:fig:depth-longer}
\end{figure}

While our goal was to improve the meta-generalization of learned optimizers to unseen wider tasks, in sections~\ref{sssec:beyond} and~\ref{sssec:beyond}, we also observed improved meta-generalization to deeper and wider networks. This discovery is entirely empirical as we did not use a parameterization that has depth transfer properties (e.g. $\mu$Depth~\citep{mudepth}). With Figure~\ref{apdx:fig:depth-longer} as evidence, we hypothesize that the reason for improved transfer to deeper models and longer training is $\mu$LOs' ability to maintain stable logits in the optimizee throughout training in contrast to SP LOs. For instance, in subfigure (a), we observe that the first layer pre-activations of depth $8$ and depth $16$ MLPs trained with LO$_M$ grow rapidly at the beginning of training, while those of deeper MLPs optimized by $\mu$LO$_M$ vary similarly to the depth-3 MLP (same depth as meta-training). In subfigure (b), we observe a similar but less drastic change in logit L1 norm as training progresses. While the L1 norm of the MLP trained by $\mu$LO$_M$ consistently grows at a stable rate throughout training, for LO$_M$ the MLP's logits undergo a change in slope after $8000$ steps of training and a near discontinuity at $13000$ steps. With the evidence we have so far, it is not possible to be certain whether the observed activation stability is the cause of the improved generalization or merely a symptom of it. That being said, these results can still help inform on favorable properties for the generalization of LOs.

\subsubsection{Evaluating the performance of $\mu$LO$_M$ on ResNet Tasks}\label{apdx:sec:explain}
\begin{figure}[h]
    \centering
    \subfloat[Train Loss: $\mu$LO$_M$]{\includegraphics[width=0.45\linewidth]{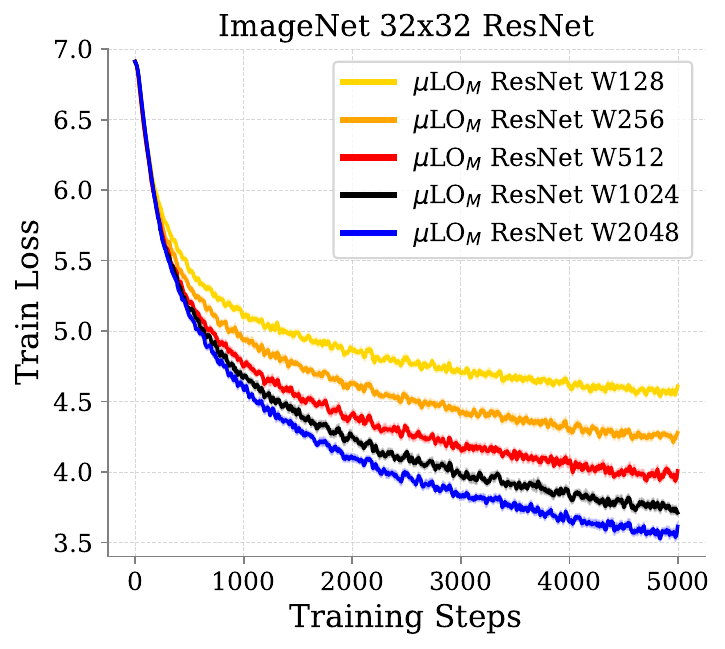}}\quad
    \subfloat[Train Loss: LO$_M$]{\includegraphics[width=0.45\linewidth]{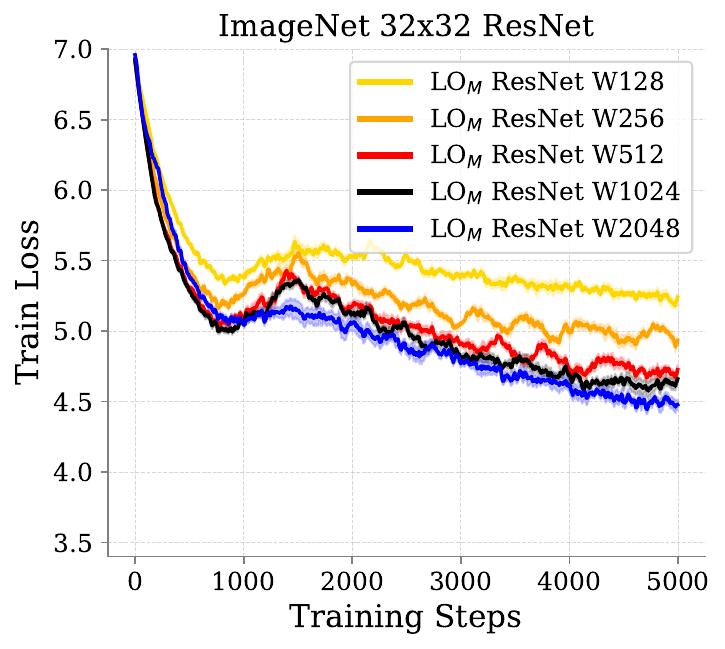}}\\
    \subfloat[Test Loss: $\mu$LO$_M$]{\includegraphics[width=0.45\linewidth]{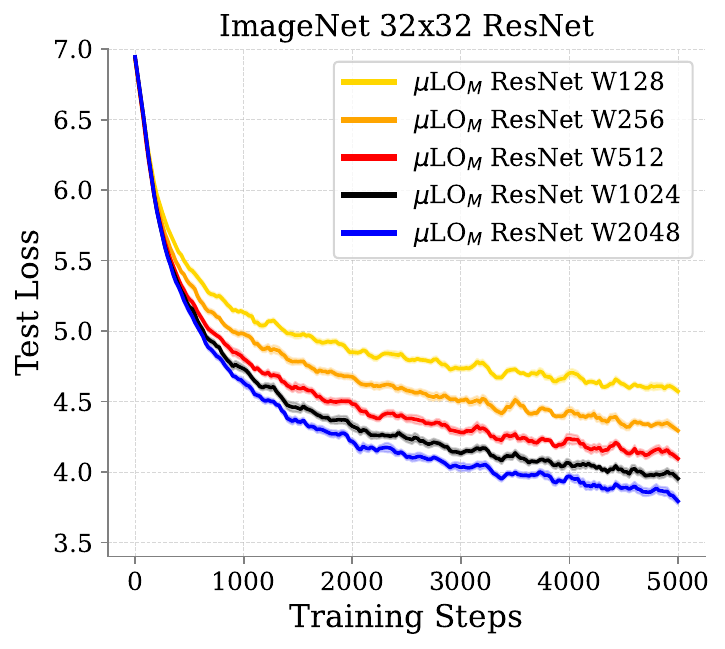}}\quad
    \subfloat[Test Loss: LO$_M$]{\includegraphics[width=0.45\linewidth]{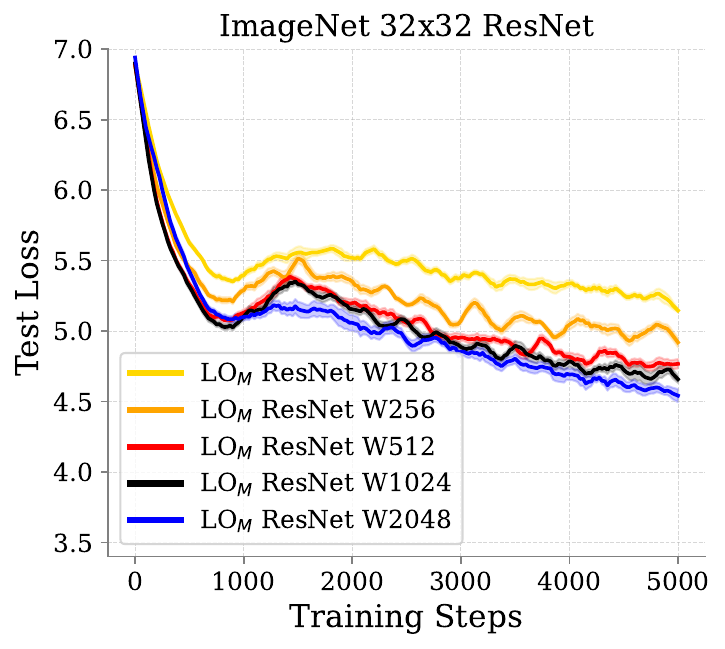}}
    \caption{\textbf{Train and Test loss across width on a ResNet task.} Each curve reports the five-seed average. We compare the meta-generalization of $\mu$LO$_M$ to LO$_M$ when optimizing ResNet tasks. We find that $\mu$LO$_M$ generalizes significantly better.}
    \label{apdx:fig:resnet}
\end{figure}

In this section, we compare the meta-generalization performance of $\mu$LO$_M$ to LO$_M$ on ResNet tasks. Figure~\ref{apdx:fig:resnet} reports train and test loss during training, while Table~\ref{apdx:tab:resnet} reports the final train and test loss across different width ResNet tasks for each optimizer. We observe that each width $\mu$LO$_M$ outperforms LO$_M$ and that the final evaluation loss for each closely tracks the training loss.

\begin{table}[h]
\centering
\caption{\textbf{Final train and test losses on ResNet tasks.} Each value reports the five-seed average and is rounded to two decimals.}
\label{apdx:tab:resnet}
\begin{tabular}{lccc}
\toprule
Optimizer & Width & Final Train Loss & Final Test Loss \\
\midrule
$\mu$LO$_M$ & 128  & $4.56 \pm 0.03$ & $4.58 \pm 0.03$ \\
$\mu$LO$_M$ & 256  & $4.25 \pm 0.02$ & $4.32 \pm 0.03$ \\
$\mu$LO$_M$ & 512  & $3.95 \pm 0.02$ & $4.11 \pm 0.02$ \\
$\mu$LO$_M$ & 1024 & $3.72 \pm 0.03$ & $3.98 \pm 0.05$ \\
$\mu$LO$_M$ & 2048 & $3.56 \pm 0.02$ & $3.83 \pm 0.03$ \\
\midrule
LO$_M$ & 128  & $5.20 \pm 0.01$ & $5.18 \pm 0.02$ \\
LO$_M$ & 256  & $4.92 \pm 0.04$ & $4.93 \pm 0.03$ \\
LO$_M$ & 512  & $4.70 \pm 0.02$ & $4.76 \pm 0.02$ \\
LO$_M$ & 1024 & $4.63 \pm 0.02$ & $4.70 \pm 0.05$ \\
LO$_M$ & 2048 & $4.47 \pm 0.03$ & $4.59 \pm 0.05$ \\
\bottomrule
\end{tabular}
\end{table}

\clearpage
\section{Coordinate evolution of MLP layers in $\mu$P for Adam and Learned Optimizers}
The following section presents the continuation of our experiments comparing pre-activation growth during training for SP LOs and $\mu$LOs with different meta-training recipes, SP adam, and $\mu$Adam.
\label{sec:apdx:activations}
\begin{figure*}[h]
    \centering
\includegraphics[width=0.99\linewidth]{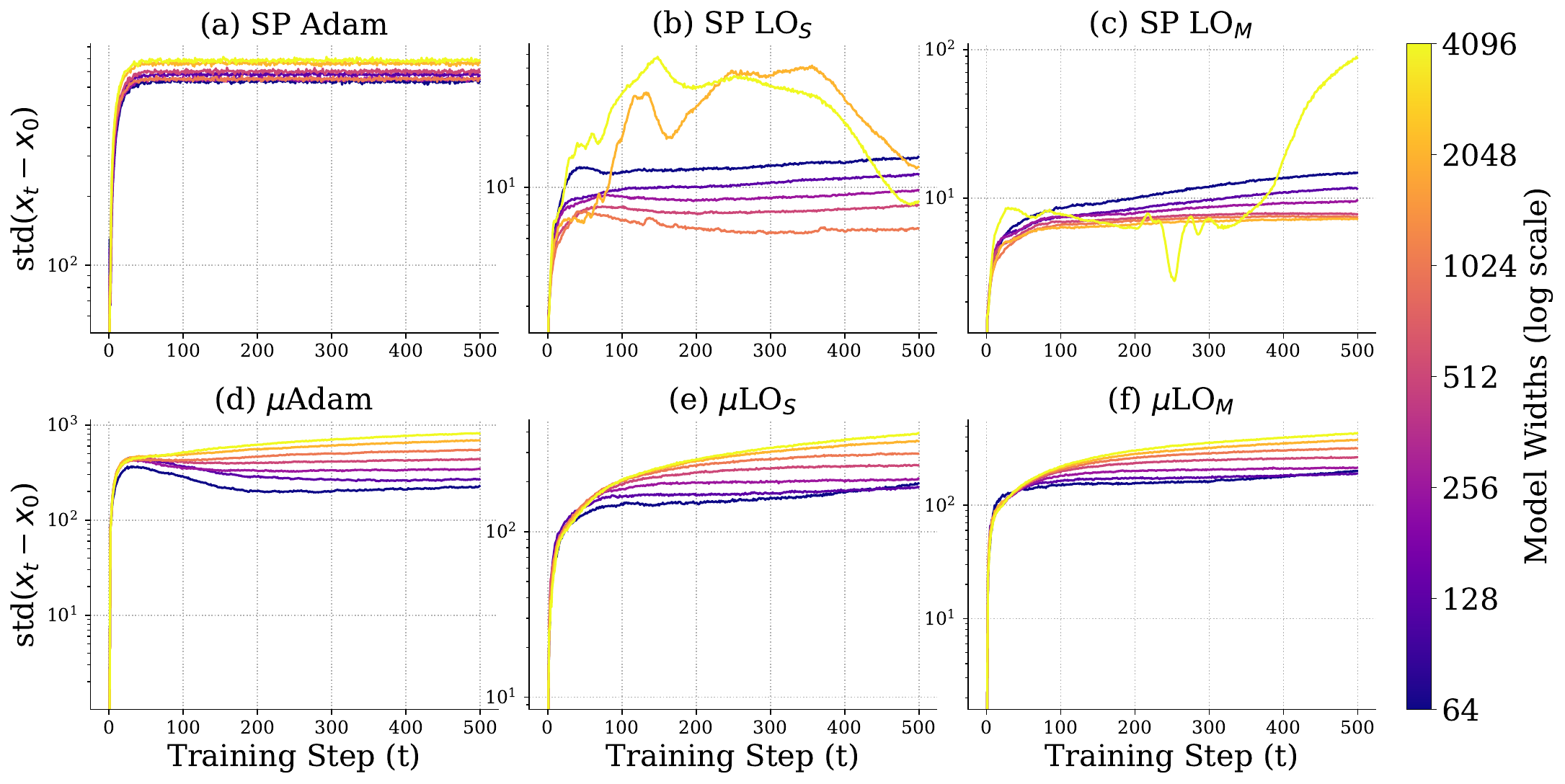}

\caption{\textbf{Layer 0 pre-activations behave harmoniously in \mup for LOs and Adam alike.} We report the evolution of coordinate-wise standard deviation between the difference of initial and current second-layer pre-activations. We observe that all models parameterized in \mup enjoy stable coordinates across widths, while the pre-activations of larger-width models in SP blow up after a number of training steps. All plots report these metrics for the first $500$ steps of a single training run.}
    \label{fig:layer0pa}
\end{figure*}

\begin{figure*}[h]
    \centering
\includegraphics[width=0.99\linewidth]{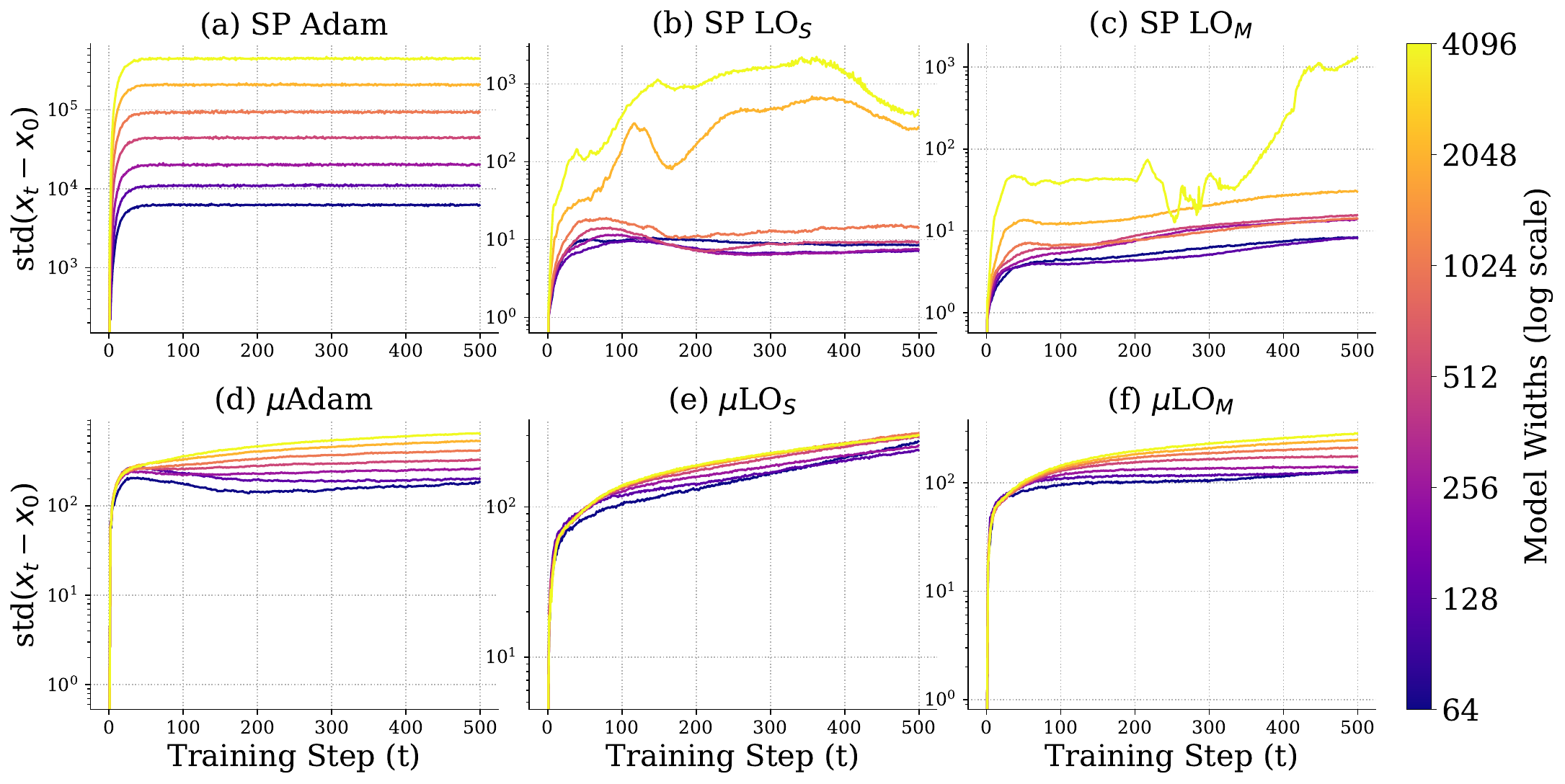}

\caption{\textbf{Layer 1 pre-activations behave harmoniously in \mup for LOs and Adam alike.} We report the evolution of coordinate-wise standard deviation between the difference of initial and current second-layer pre-activations. We observe that all models parameterized in \mup enjoy stable coordinates across widths, while the pre-activations of larger-width models in SP blow up after a number of training steps. All plots report these metrics for the first $500$ steps of a single training run.}
    \label{fig:layer1pa}
\end{figure*}



\begin{figure*}[h]
    \centering
\includegraphics[width=0.99\linewidth]{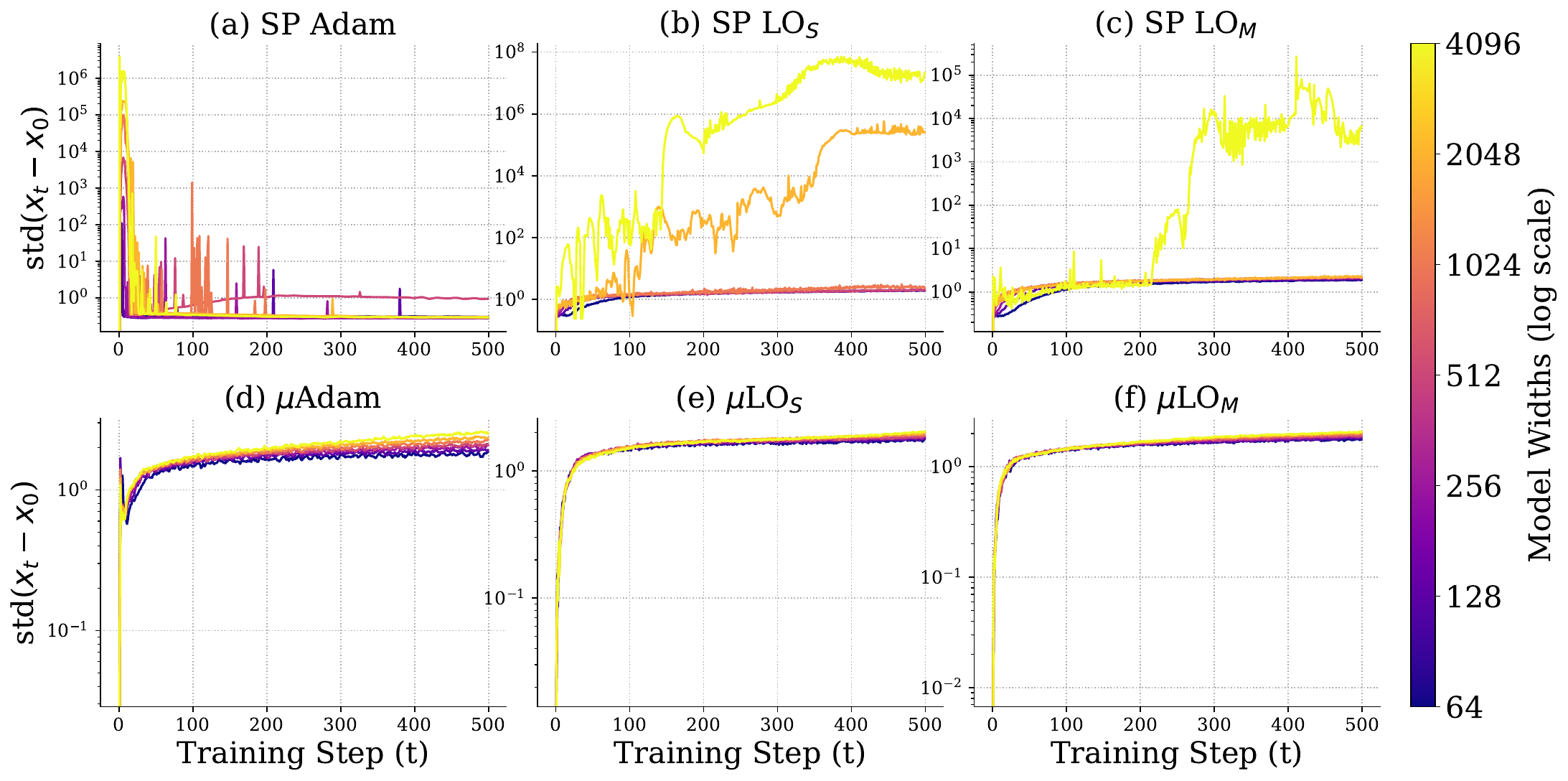}

\caption{\textbf{Logits behave harmoniously in \mup for LOs and Adam alike.} We report the evolution of coordinate-wise standard deviation between the difference of initial and current second-layer pre-activations. We observe that all models parameterized in \mup enjoy stable logits across widths, while the pre-activations of larger-width models in SP blow up after a number of training steps. All plots report these metrics for the first $500$ steps of a single training run.}
    \label{fig:logits}
\end{figure*}
\clearpage

\end{document}

%% file: math_commands.tex
\usepackage{amsmath,amsfonts,bm}

\def\eqref#1{equation~\ref{#1}}

\def\1{\bm{1}}

\def\vphi{{\bm{\phi}}}

\def\vtheta{{\bm{\theta}}}

\def\vb{{\bm{b}}}
\def\vc{{\bm{c}}}

\def\vr{{\bm{r}}}

\def\vu{{\bm{u}}}
\def\vv{{\bm{v}}}
\def\vw{{\bm{w}}}
\def\vx{{\bm{x}}}

\def\mA{{\bm{A}}}

\def\mF{{\bm{F}}}

\def\mH{{\bm{H}}}

\def\mL{{\bm{L}}}
\def\mM{{\bm{M}}}

\def\mR{{\bm{R}}}

\def\mT{{\bm{T}}}
\def\mU{{\bm{U}}}
\def\mV{{\bm{V}}}
\def\mW{{\bm{W}}}

\DeclareMathAlphabet{\mathsfit}{\encodingdefault}{\sfdefault}{m}{sl}
\SetMathAlphabet{\mathsfit}{bold}{\encodingdefault}{\sfdefault}{bx}{n}

\def\gD{{\mathcal{D}}}

\def\gL{{\mathcal{L}}}

\def\gT{{\mathcal{T}}}

\newcommand{\E}{\mathbb{E}}

\newcommand{\R}{\mathbb{R}}

